\documentclass[journal]{IEEEtran}

\usepackage{graphics} 
\usepackage{amsmath} 
\usepackage[space]{cite}
\usepackage{subfigure}
\usepackage{tabulary}
\usepackage{multirow}
\usepackage{caption}
\usepackage{todonotes}
\graphicspath{{figs/}}
\usepackage[colorlinks,linkcolor=blue,anchorcolor=blue,citecolor=red]{hyperref} 
\usepackage{wrapfig,lipsum,booktabs}
\usepackage{color}
\usepackage{algorithmic} 
\usepackage[ruled,vlined,commentsnumbered]{algorithm2e}

\usepackage[flushleft]{threeparttable}
\usepackage{gensymb}  
\usepackage{amsmath}
\usepackage{amssymb}

\usepackage{flushend}

\graphicspath{{figs/}}

\usepackage{siunitx}
\usepackage{url}   

\begin{document}

\title{\LARGE \bf
CubeSLAM: Monocular 3D Object SLAM
}

\author{Shichao Yang, Sebastian Scherer 

\thanks{The authors are with the Robotics Institute, Carnegie Mellon University, Pittsburgh, PA, USA. Email of first author: \{shichaoy@andrew.cmu.edu, 2013ysc@gmail.com\}; Second author: basti@andrew.cmu.edu}
\thanks{The work was supported by the Amazon Research Award \#2D-01038138.}
\thanks{Digital Object Identifier 10.1109/TRO.2019.2909168}

}

\maketitle

\begin{abstract}
We present a method for single image 3D cuboid object detection and multi-view object SLAM in both static and dynamic environments, and demonstrate that the two parts can improve each other. Firstly for single image object detection, we generate high-quality cuboid proposals from 2D bounding boxes and vanishing points sampling. The proposals are further scored and selected based on the alignment with image edges. Secondly, multi-view bundle adjustment with new object measurements is proposed to jointly optimize poses of cameras, objects and points. Objects can provide long-range geometric and scale constraints to improve camera pose estimation and reduce monocular drift. Instead of treating dynamic regions as outliers, we utilize object representation and motion model constraints to improve the camera pose estimation. The 3D detection experiments on SUN RGBD and KITTI show better accuracy and robustness over existing approaches. On the public TUM, KITTI odometry and our own collected datasets, our SLAM method achieves the state-of-the-art monocular camera pose estimation and at the same time, improves the 3D object detection accuracy.



\end{abstract}

\begin{IEEEkeywords}
Simultaneous localization and mapping (SLAM), object detection, dynamic SLAM, object SLAM
\end{IEEEkeywords}

\IEEEpeerreviewmaketitle

\section{Introduction}
\label{sec:object intro}

\IEEEPARstart{O}{bject} detection and Simultaneous localization and mapping (SLAM) are two important tasks in computer vision and robotics. For example, in autonomous driving, vehicles need to be detected in 3D space in order to remain safe. In augmented reality, 3D objects also need to be localized for more realistic physical interactions. Different sensors can be used for these tasks such as laser-range finders, stereo or RGB-D cameras, which can directly provide depth measurement. Alternatively, monocular cameras are attractive because of their low cost and small size. Most of the existing monocular approaches solve object detection and SLAM separately while also depend on prior object models which limit their application to general environments. Therefore, we focus on the general 3D object mapping problem by solving object detection and multi-view object SLAM jointly in both static and dynamic environments.

\begin{figure}[t]
  \centering
   \subfigure[]{\includegraphics[scale=0.16]{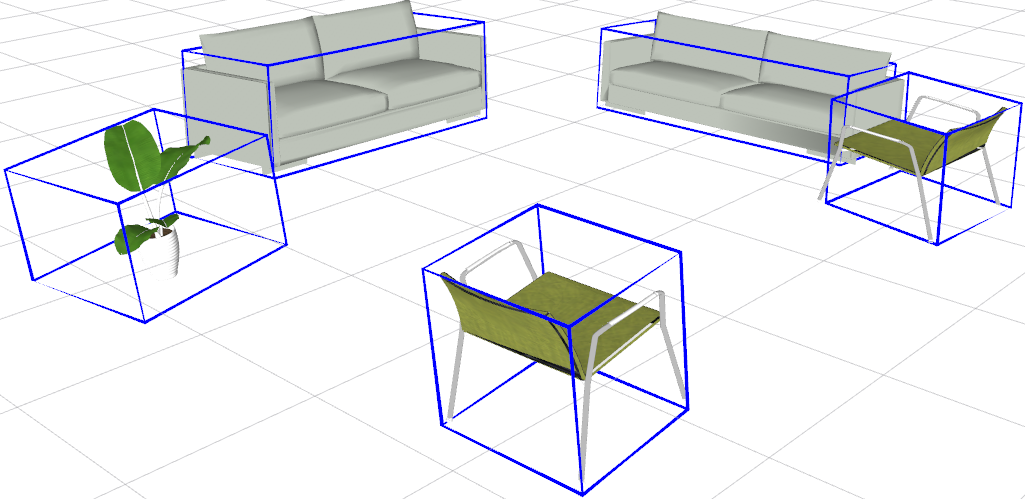} \label{fig:icl mapping}}
   \subfigure[]{\includegraphics[scale=0.27]{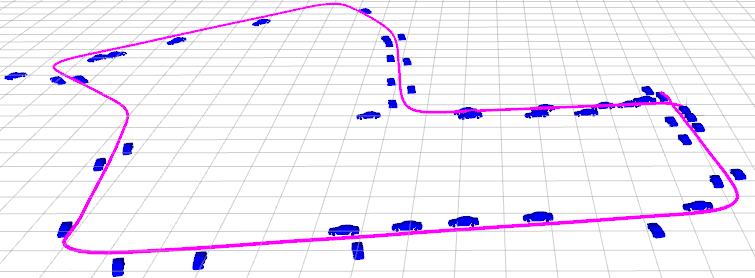} \label{fig:kitti mapping}}
   \caption{ Monocular 3D object detection and mapping without requiring prior object models. Mesh model is just for visualization and not used for detection. (a) ICL NUIM data with various objects, whose position, orientation and dimension are optimized by SLAM. (b) KITTI 07. With object constraints, our monocular SLAM can build a consistent map and reduce scale drift, without loop closure and constant camera height assumption.}
\end{figure}

For object detection, many algorithms are able to detect different 2D objects with various size and viewpoints in large datasets using convolutional neural networks (CNNs) \cite{he2017mask}. 3D object objection is more challenging and has also attracted attention recently, such as for vehicle detection \cite{xiang2017subcategory,chabot2017deep}. For SLAM or Structure from Motion (SfM), the classic approach is to track visual geometric features such as points \cite{mur2015orb}, lines \cite{chao2017edge}, planes \cite{kaess2015simultaneous} across frames then minimize the reprojection or photometric error through bundle adjustment (BA). However, apart from these low-level features in the environments, objects are also important components which have not been well explored in SLAM. Detecting and mapping 3D objects can greatly improve robot intelligence for environment understanding and human-robot interaction. In addition, objects used as SLAM landmarks can also provide additional semantic and geometric constraints to improve camera pose estimation. 

Most existing SLAM approaches assume the environment to be static or mostly static. Features from dynamic regions are often treated as outliers and not used for camera pose estimation \cite{mur2015orb}, however this assumption may not hold in many practical environments. For example, there are many moving vehicles and pedestrians on the road. It is also an important task to detect and predict the trajectory of moving objects in many applications.



In this work, we propose a system to combine 2D and 3D object detection with SLAM pose estimation together for both static and dynamic environments. Given the detected 2D object, many 3D cuboid proposals are efficiently generated through vanishing point (VP) sampling, under the assumptions that the cuboid will fit 2D bounding box tightly after projection. Then the selected cuboid proposals are further optimized with points and cameras through multi-view BA. Objects are utilized in two ways: to provide geometry and scale constraints in BA, and to provide depth initialization for points difficult to triangulate. The estimated camera poses from SLAM are also used for single-view object detection. Lastly in the dynamic case, instead of treating moving objects as outliers, we jointly optimize the trajectories of camera and objects based on dynamic point observation and motion model constraints. In summary, our contributions are as follows:


\begin{itemize}
\item An efficient, accurate and robust single image 3D cuboid detection approach without requiring prior object models.
\item An object SLAM method with novel measurements between cameras, objects and points, that achieves better pose estimation on many datasets including KITTI benchmark.
\item Results demonstrating that object detection and SLAM benefit each other.
\item A method to utilize moving objects to improve pose estimation in dynamic scenarios.
\end{itemize}

In the following, we first introduce the single image detection in Sec. \ref{sec:object detect}, then explain the object SLAM in static (Sec. \ref{sec:object slam}) and dynamic environments (Sec. \ref{sec:dynamic slam}), followed by implementations in Sec. \ref{sec:implementation} and experiments in Sec. \ref{sec:experiment single view} and Sec. \ref{sec:experiment object SLAM}. Part of the code is public available \footnote{Some code available at \url{https://github.com/shichaoy/cube_slam}}.





\section{Related Work}
\label{sec:object review}



\subsection{Single image 3D object detection}




3D object detection from a single image is much more challenging compared to 2D because more object pose variables and the camera projective geometry need to be considered. Existing 3D detection approaches can be divided into two categories: with or without shape priors, such as CAD models. With prior models, the best object pose to align with RGB images can be found through keypoint Perspective n-Point (PnP) matching \cite{murthy2017reconstructing}, hand-crafted texture features \cite{lim2013parsing} or more recent deep networks \cite{chabot2017deep,kehl2017ssd}. 

Without prior models, objects are usually represented by cuboids. The typical approach is to combine geometry modelling with learning. For example, objects can be generated by a combination of Manhattan edges or rays through VPs \cite{xiao2012localizing,hedau2010thinking}. Chen \textit{et al.} proposed to exhaustively sample many 3D boxes on the ground then select based on various context features \cite{chen2016monocular}. Two similar work to us is \cite{mousavian20163d,li2018stereo} which used projective geometry to find cuboids to fit the 2D bounding box tightly. We extend it to work without prediction of object size and orientation.





\subsection{Multi-view object SLAM}

There are many point-based visual SLAM algorithms such as ORB SLAM \cite{mur2015orb}, and DSO \cite{engel2017direct}, which can achieve impressive results in general environments. Object-augmented mapping is also explored in recent years. There are typically two categories of them, either decoupled or coupled. The decoupled approaches first build a SLAM point cloud map then further detect and optimize 3D object poses based on point cloud clustering and semantic information \cite{dong2017visual, pillai2015monocular, dame2013dense}. It showed improved results compared to 2D object detections, however it didn't change the SLAM part, thus the decoupled approach may fail if SLAM cannot build a high quality map.

The coupled approach is usually called object-level SLAM.  Bao \textit{et al.} proposed the first Semantic SfM to jointly optimize camera poses, objects, points and planes \cite{bao2012semantic}. Salas \textit{et al.} \cite{salas2013slam++} proposed a practical SLAM system called SLAM++ using RGB-D cameras and prior object models. Frost \textit{et al.} represented objects as spheres to correct the scale drift of monocular SLAM \cite{frost2018recovering}, similarly in \cite{sucar2017bayesian}. Recently, a real time monocular object SLAM using the prior object models was proposed in \cite{galvez2016real}. Rubino \textit{et al.} \cite{rubino20183d} solved multi-view 3D ellipsoid object localization analytically and QuadriSLAM \cite{nicholson2018quadricslam} extended it to an online SLAM without prior models. Uncertain data association of object SLAM is addressed in \cite{bowman2017probabilistic}. Yang \textit{et al.} \cite{syang2016popslam} proposed a similar idea to combine scene understanding with SLAM but only applied to planes.

Recently, there is also some end-to-end deep learning-based SLAM without object representations, such as DVSO\cite{yang2018deep}, DeepVO\cite{wang2017deepvo}. They have achieved great performance on KITTI datasets, however, it is still unclear if they would generalize to novel environments.

\subsection{Dynamic environment SLAM}

SLAM in dynamic environments has been a challenging problem. Most existing approaches treat dynamic region features as outliers and only utilize static background for pose estimation \cite{mur2015orb,barsan2018robust,bescos2018dynslam}. After the static SLAM problem is solved, some other works additionally detect, track, and optimize the trajectory of dynamic objects in order to build a complete 3D map \cite{li2018stereo,song2015joint,reddy2015dynamic}. The optimization is based on the object's reprojection error, object motion model, and also the point feature observations on the object. However, these approaches are likely to fail in highly dynamic environments due to the lack of reliable static background features.

There is a recent work utilizing dynamic point BA to improve camera pose estimation, based on the rigid shape and constant motion assumption \cite{henein18exploiting}, however, the paper showed limited real dataset results and didn't explicitly represent objects in the map.





\section{Single Image 3D Object Detection}
\label{sec:object detect}

\subsection{3D box proposal generation}
\label{sec:proposal generation}

\subsubsection{Principles}
\label{sec:obj detect core}

Instead of randomly sampling object proposals in 3D space, we utilize the 2D bounding box to efficiently generate 3D cuboid proposals. A general 3D cuboid can be represented by 9 DoF parameters: 3 DoF position $\mathbf{t}=[t_x,t_y,t_z]$, 3 DoF rotation $R$ and 3 DoF dimension $\mathbf{d}=[d_x,d_y,d_z]$. The cuboid coordinate frame is built at the cuboid center, aligned with the main axes. The camera intrinsic calibration matrix $K$ is also known. Based on the assumptions that the cuboid's projected corners should fit the 2D bounding box tightly, there are only four constraints corresponding to four sides of the 2D box, therefore, it is not possible fully constrain all 9 parameters. Other information is needed for example the provided or predicted object dimensions and orientations used in many vehicle detection algorithms \cite{chen2016monocular,mousavian20163d,li2018stereo}. Rather than relying on the predicted dimensions, we utilize the VP to change and reduce the regression parameters in order to work for general objects.

The VP is the parallel lines' intersection after projection onto perspective images \cite{hartley2003multiple}. A 3D cuboid has three orthogonal axes and can form three VPs after projections depending on object rotation $R$ \textit{wrt.} camera frame and calibration matrix $K$:

\begin{equation}
\text{VP}_i =K R_{col(i)}, \quad i \in \left\lbrace 1,2,3 \right\rbrace
\label{eq:vp formula}
\end{equation}
where $R_{col(i)}$ is the $i^{th}$ column of $R$.

\subsubsection{Get 2D corners from the VP}
\label{sec:2d corner from vp}
We first show how to get eight 2D cuboid corners based on the VP. Since at most three cuboid faces can be observed simultaneously, we can divide the cuboid configurations into three common categories based on the number of observable faces shown in Fig. \ref{fig:cube_proposal}. Each configuration can be left-right symmetric. Here we explain Fig. \ref{fig:cube_proposal}(a) in more detail. Suppose three VPs and top corner $p_1$ are known or estimated, and $\times$ represents the intersection of two lines, then $p_2 = \overline{(\text{VP}_1, p_1)} \times \overline{(B, C)}$, similarly for $p_4$.  $p_3 = \overline{(\text{VP}_1, p_4)} \times \overline{(\text{VP}_2, p_2)}, p_5 = \overline{(\text{VP}_3, p_3)} \times \overline{(C,D)}$, similarly for the remaining corners.

\subsubsection{Get 3D box pose from 2D corners}
\label{sec:3d pose from 2d}
After we get the cuboid corners in 2D image space, we need to estimate the cuboid's 3D corners and pose. We divide the objects into two scenarios.

\begin{figure}[t]
  \vspace{0.5em}
  \centering
   \includegraphics[trim={0cm 9.1cm 10cm 0cm},clip,scale=0.37]{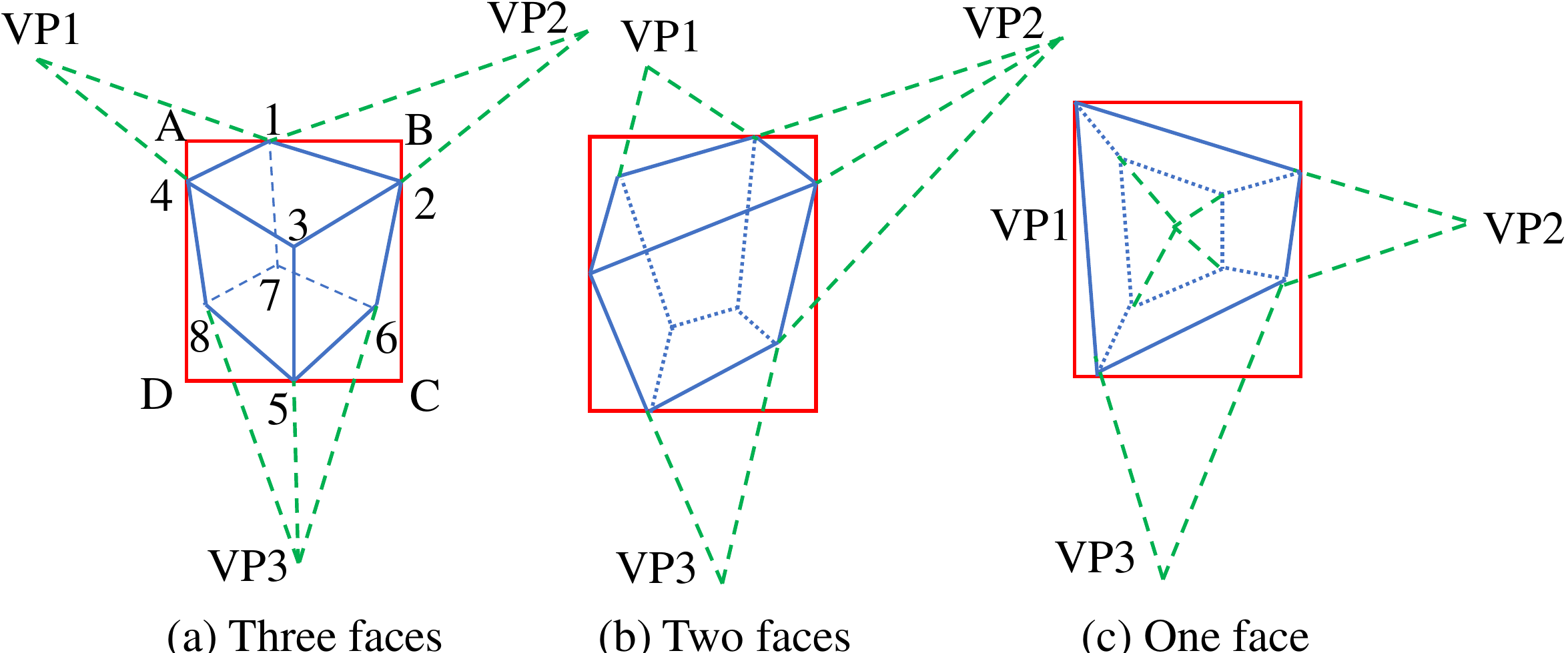}
   \caption{Cuboid proposals generation from 2D object box. If vanishing points and one corner are estimated, the other seven corners can also be computed analytically. For example in (a), given corner 1, corner 2 and 4 can be determined through line intersection, and same as other corners.
   }
   \label{fig:cube_proposal}
\end{figure}

Arbitrary pose objects:
We use PnP solver to solve the general cuboid's 3D position and dimensions up to a scale factor due to the monocular scale ambiguity. Mathematically, the cuboid's eight 3D corners in the object frame are $\left[\pm d_x, \pm d_y, \pm d_z \right]/2$ and in the camera frame are: $R\left[ \pm d_x, \pm d_y, \pm d_z \right]/2 +\mathbf{t}$. As shown in Fig. \ref{fig:cube_proposal}(a), we can select four adjacent corners such as 1,2,4,7, which can be projected from the above 3D corners for example corner 1:
\begin{equation}
\label{eq:corner pnp}
p_1 = \pi \left(R\left[d_x, d_y, d_z \right]/2 +\mathbf{t}) \right)
\end{equation}
\noindent $\pi$ is the camera projective function and $p_i \ (i=1,2...8)$ is one of the eight 2D object corners. Each corner provides two constraints thus four corners can fully constrain the object pose (9 Dof) except the scale. Iterative or non-iterative PnP solvers can be found in \cite{hartley2003multiple}.

Ground objects:
For ground objects lying on the ground plane, we can further simplify the above the process and get the scale factor more easily. We build the world frame on the ground plane then object's roll/pitch angle is zero. Similar to the previous section, we can get eight 2D corners from VP. Then instead of using the complicated PnP solver in Equation \ref{eq:corner pnp}, we can directly back-project ground corner pixels to the 3D ground plane and subsequently compute other vertical corners to form a 3D cuboid. This approach is very efficient and has analytical expressions. For example for corner 5 on the 3D ground plane expressed by $[\mathbf{n},m]$ (normal and distance in camera frame), the corresponding 3D corner $\mathbf{P}_5$ is the intersection of backprojected ray $K^{-1}p_5$ with the ground plane:
\begin{equation}
\label{eq:projection equation}
\mathbf{P}_5=\frac{-m}{\mathbf{n}^\top(K^{-1}p_5)}K^{-1}p_5
\end{equation}

\noindent Similarly, a more detailed projection process is explained in the \cite{syang2016popslam}. The scale is determined by the camera height in the projection process.


\subsubsection{Sample VP and Summary}
\label{sec:object sampling}

From the analysis in the previous two sections, the box estimation problem changes to how to get three VPs and one top 2D corner, because after we get the VP, we can use Sec. \ref{sec:2d corner from vp} to compute 2D corners, then use Sec. \ref{sec:3d pose from 2d} to compute 3D box.

From Equation \ref{eq:vp formula}, VP is determined by object rotation matrix $R$. Though deep networks can be used to directly predict them with large amounts of data training, we choose to sample them manually then score (rank) them for the purpose of generalizability.

For general objects, we need to sample the full rotation matrix $R$, however for ground objects, camera's roll/pitch and object's yaw are used and sampled to compute $R$. More importantly, in datasets such as SUN RGBD or KITTI, camera roll/pitch are already provided. For mutli-view video data, we use SLAM to estimate camera poses. Therefore the sampling space is greatly reduced and also becomes more accurate. In this paper's experiments, we only consider ground objects.

\subsection{Proposal scoring}
\label{sec:proposal scoring}
\begin{figure}[t]
  \centering
   \includegraphics[scale=0.30]{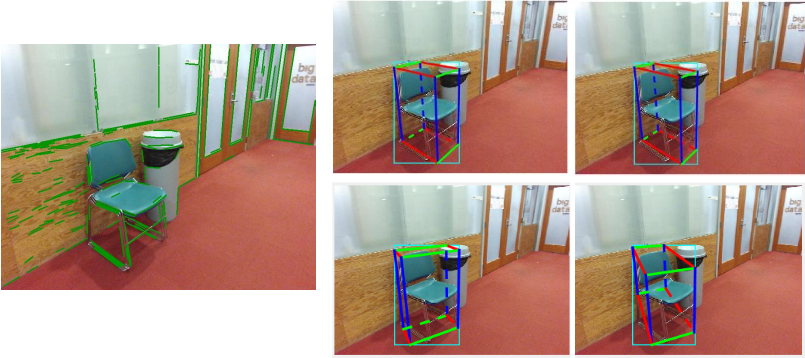}
   \caption{Cuboid proposal scoring. (\textbf{Left}) Edges used to score the proposals. (\textbf{Right}) Cuboid proposals generated from the same 2D cyan bounding box. The top left is the best and bottom right is the worst after scoring.}
   \label{fig:cube score}
\end{figure}

After sampling many cuboid proposals, we define cost functions to score them as shown in Fig. \ref{fig:cube score}. Different functions have been proposed such as semantic segmentation \cite{chen2016monocular}, edge distance \cite{lim2013parsing}, HOG features \cite{xiao2012localizing}. We propose some fast and effective cost functions to align the cuboid with image edge features. This approach works best for ``boxy with clear edge" objects, but also works decently well for bicycles and toilets \textit{etc.} as shown in later experiments due to constraints from VP and robust edge filtering. We first denote the image as $I$ and cuboid proposal as $O=\lbrace R,\mathbf{t},\mathbf{d} \rbrace $ defined in Sec. \ref{sec:obj detect core}, then the cost function is defined as:


\begin{equation}
E(O|I) = \phi_{dist} (O,I) + w_1 \phi_{angle} (O,I) + w_2 \phi_{shape} (O)
\label{eq:score proposal}
\end{equation}
where $\phi_{dist}, \phi_{angle}, \phi_{shape}$ are three kinds of costs which will be explained as follows. $w_1$ and $w_2$ are weight parameters between different costs. We set $w_1 = 0.8, w_2 = 1.5$ after manual search on small sample datasets.

\subsubsection{Distance error $\phi_{dist}(O,I)$}
The 2D cuboid edges should match with the actual image edges. We first detect Canny edges and build a distance transform map based on them. Then for each visible cuboid edge (solid blue line in Fig. \ref{fig:cube_proposal}(a)), we evenly sample 10 points on it and summarize all the distance map value divided by the 2D box's diagonal. This is similar to the Chamfer distance \cite{xiao2012localizing}.

\subsubsection{Angle alignment error $\phi_{angle}(O,I)$}
The distance error is sensitive to the noisy false positive edges such as object surface textures. Therefore, we also detect long line segments \cite{von2008lsd} (shown as the green lines in Fig. \ref{fig:cube score}) and measure whether their angles align with vanishing points. These lines are first associated with one of three VPs based on the point-line support relationship \cite{hedau2010thinking}. Then for each $\text{VP}_i$, we can find two outmost line segments with the smallest and largest slope denoted as $\langle l_{i\_ms},l_{i\_mt} \rangle$ and $\langle l_{i\_ns},l_{i\_nt} \rangle$ respectively. $\langle a,b \rangle$ represents the slope angle of a line with two endpoints $a,b$. Finally the angle alignment error is:

\begin{equation}
\begin{split}
\phi_{angle}(O,I) = \sum_{i=1:3} & \| \langle l_{i\_ms},l_{i\_mt} \rangle - \langle \text{VP}_i,l_{i\_mt}\rangle\| + \\
&\|\langle l_{i\_ns},l_{i\_nt} \rangle - \langle \text{VP}_i,l_{i\_nt} \rangle\|
\end{split}
\end{equation}



\subsubsection{Shape error $\phi_{shape}(O)$}
The previous two costs can be evaluated efficiently just in 2D image space. However, similar 2D cuboid corners might generate quite different 3D cuboids. We add a cost to penalize the cuboids with a large skew (length/width) ratio. More strict priors could be applied for example the estimated or fixed dimensions of specific types of objects.

\begin{equation}
\phi_{shape}(O) = \max(s-\sigma,0)
\end{equation}
where $\ s=\max(d_x/d_y,d_y/d_x)$ is the object skew ratio and $\sigma$ is a threshold set be $1$ in our experiments. If $s<\sigma$, no penalty is applied.

\section{Object SLAM}
\label{sec:object slam}
We extend the single image 3D object detection to multi-view object SLAM to jointly optimize object pose and camera pose. The system is built on feature point-based ORB SLAM2 \cite{mur2015orb}, which includes the front-end of camera tracking and back-end of BA. Our main change is the modified BA to include objects, points and camera poses together, which will be explained in detail in this section. Other SLAM implementation details are in Sec. \ref{sec:SLAM implementation}. Static objects are used in this section and dynamic objects are addressed in the next section.




\subsection{Bundle Adjustment Formulation}
\label{sec:slam pipeline}

BA is the process to jointly optimize different map components such as camera poses and points \cite{mur2015orb} \cite{engel2017direct}.  Points are also used in most of our experiments because objects alone usually cannot fully constrain camera poses. 
If we denote the set of camera poses, 3D cuboids and points as $C=\{C_i\}, O=\{O_j\}, P=\{P_k\}$ respectively, BA can be formulated as a nonlinear least squares optimization problem:


\begin{equation}
\begin{split}
C^*,O^*,P^* = & \underset{\{C,O,P\}}{\arg \min} \sum_{C_i,O_j,P_k} \parallel \mathbf{e}(c_i,o_j) \parallel_{\Sigma_{ij}}^2 +
\\
&\parallel \mathbf{e}(c_i,p_k) \parallel_{\Sigma_{ik}}^2 + \parallel \mathbf{e}(o_j,p_k) \parallel_{\Sigma_{jk}}^2
\end{split}
\end{equation}

\noindent where $\mathbf{e}(c,o),\mathbf{e}(c,p),\mathbf{e}(o,p)$ represents the measurement error of camera-object, camera-point, object-point respectively. $\Sigma$ is covariance matrix of different error measurements. Definitions of variables and errors are in the following. Then the optimization problem can be solved by Gauss-newton or Levenberg-Marquardt algorithm available in many libraries such as g2o \cite{kummerle2011g} and iSAM \cite{kaess2008isam}.

\vspace{0.5em}
\noindent \textit{Notations}: Camera poses are represented by $T_c \in SE(3)$ and points are represented by $P \in \mathbb{R}^3$. As explained in Section \ref{sec:obj detect core}, cuboid objects are modelled as 9 DoF parameters: $O=\{T_o, \mathbf{d}\}$ where $T_o =[R \ \mathbf{t}] \in SE(3)$ is 6 DoF pose, and $\mathbf{d} \in \mathbb{R}^3$ is the cuboid dimension. In some environments such as KITTI, we can also use the provided object dimension then $\mathbf{d}$ is not needed to optimize. Subscript $m$ indicates the measurement. The coordinate system is shown in Fig. \ref{fig:slam measurement}.

\subsection{Measurement Errors}

\subsubsection{Camera-Object measurement}
\label{sec:obj cam meas}

We propose two kinds of measurement errors between objects and cameras.

\paragraph{3D measurements} The first is 3D measurement utilized when the 3D object detection is accurate for example if a RGBD sensor is used. The detected object pose $O_m = (T_{om}, \mathbf{d}_m)$ from single image detection in Section \ref{sec:proposal generation} serves as the object measurement from the camera frame.  To compute its measurement error, we transform the landmark object to the camera frame then compare with the measurement:

\begin{equation}
\label{eq:object camera}
e_{co\_3D} = \lbrack \log \big( (T_c^{-1}T_o)T_{om}^{-1} \big)^{\vee}_{\mathfrak{se}_3} \quad \mathbf{d}-\mathbf{d}_m \rbrack
\end{equation}
where $\log$ maps the $SE3$ error into 6 DoF tangent vector space, therefore $e_{co\_3D} \in \mathbb{R}^9$. Huber robust cost function is applied to all measurement errors to improve the robustness \cite{mur2015orb}.

We need to note that, without prior object models, our image-based cuboid detection cannot differentiate between the front or back of objects. For example, we can represent the same cuboid by rotating the object coordinate frame by $90 \degree$ and swapping length with width value. Therefore, we need to rotate along the height direction for ${0, \pm 90 \degree, 180 \degree}$ to find the smallest error in Eq. \ref{eq:object camera}.

\begin{figure}[t]
  \centering
   \subfigure[]{\includegraphics[trim={0cm 10.2cm 17cm 0cm},clip,scale=0.43]{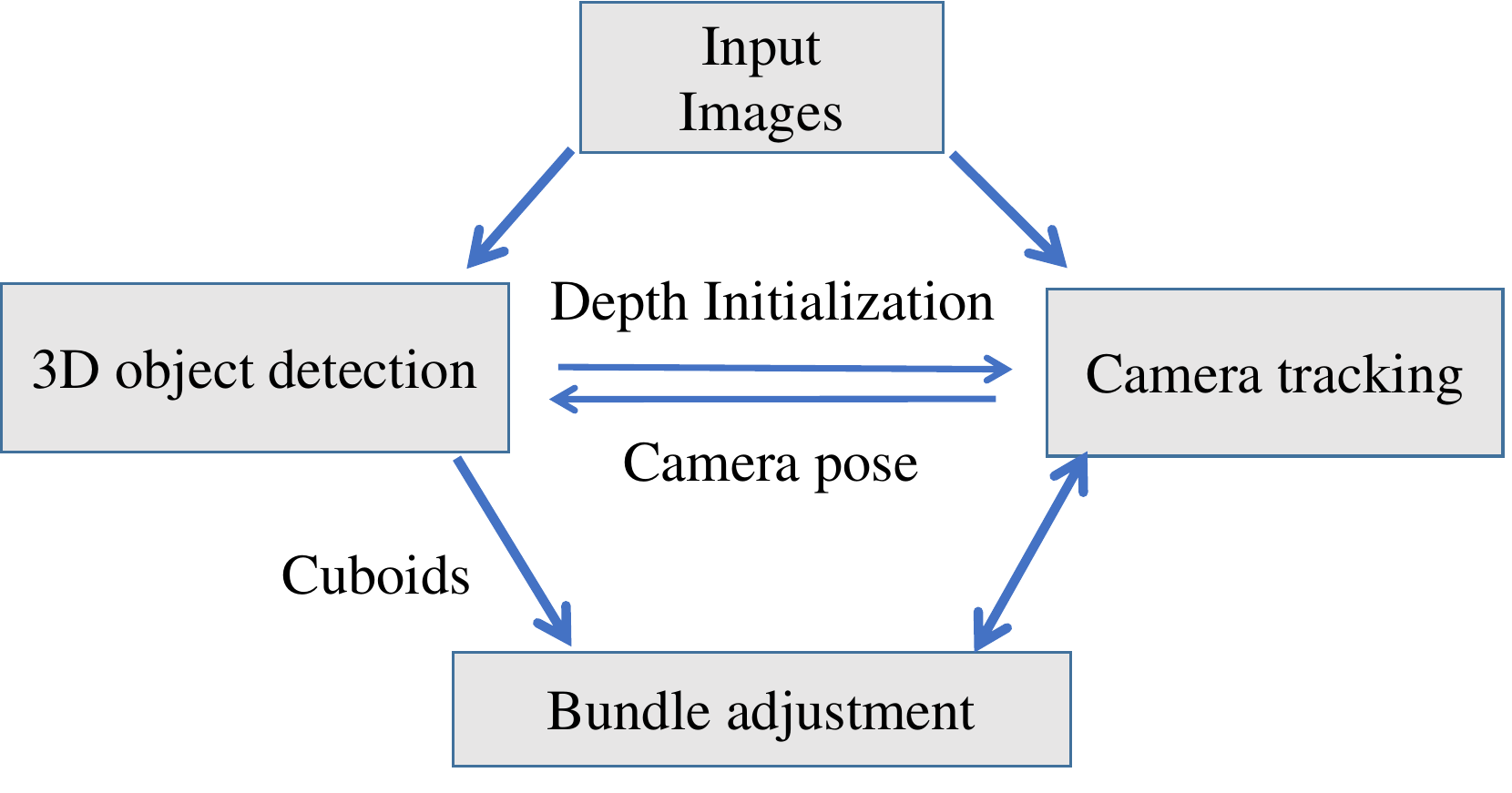} \label{fig:slam flow} }
   \subfigure[]{\includegraphics[scale=0.25]{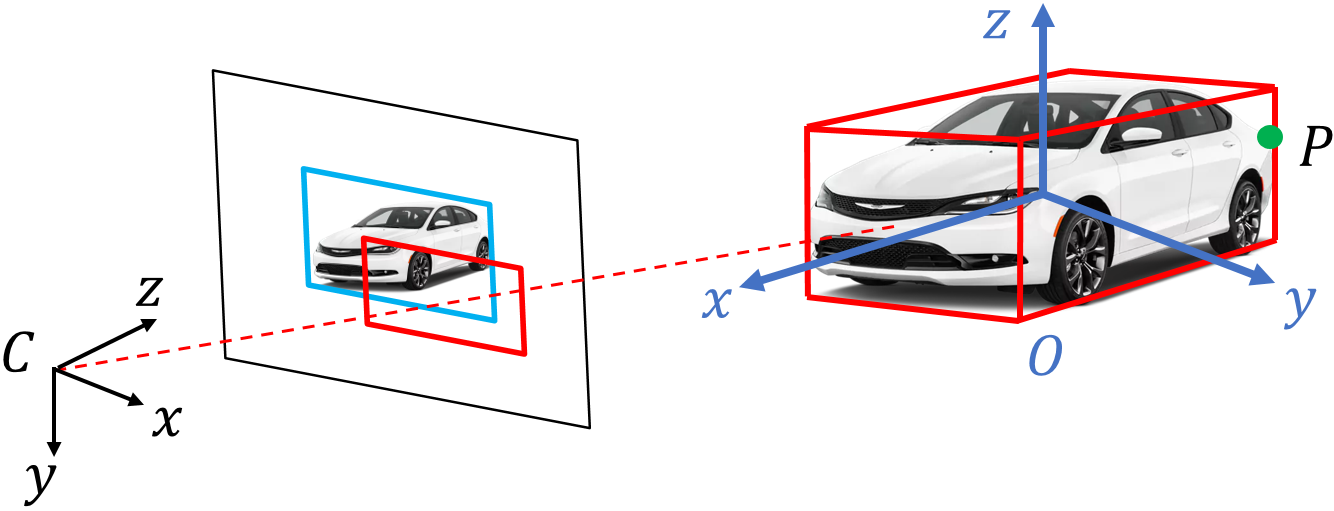} \label{fig:slam measurement}}
   \caption{(a) Our object SLAM pipeline. Single view object detection provides cuboid landmark and depth initialization for SLAM while SLAM can estimate camera pose for more accurate object detection. (b) Showing the coordinate system and measurement errors between cameras, objects and points during BA.}
   \label{fig:flow char}
\end{figure}

\paragraph{2D measurements}
For the 2D measurement, we project the cuboid landmark onto the image plane to get the 2D bounding box shown as the red rectangle in Fig. \ref{fig:slam measurement} then compare it with the blue detected 2D bounding box. In more detail, we project the eight corners onto the image and find the min and max of the projected pixels' x y coordinates to form a rectangle:

\begin{equation}
\begin{split}
[u,v]_{min} &= \min \lbrace \pi \left(R\left[\pm d_x, \pm d_y, \pm d_z \right]/2 +\mathbf{t}) \right) \rbrace  \\
[u,v]_{max} &= \max \lbrace \pi \left(R\left[\pm d_x, \pm d_y, \pm d_z \right]/2 +\mathbf{t}) \right) \rbrace  \\
\mathbf{c} &=  ([u,v]_{min}+[u,v]_{max})/2 \\
\mathbf{s} &=  [u,v]_{max} - [u,v]_{min}
\end{split}
\end{equation}

\noindent where $[u,v]_{min,max}$ are the min/max xy coordinates of all eight projected corners, namely the top left and bottom right corners of projected rectangle. $\mathbf{c}$ and $\mathbf{s}$ is the center and size of the 2D box. Both of them are two dimensional vectors therefore $[\mathbf{c},\mathbf{s}] \in \mathbb{R}^4$. The 4D rectangle error is then defined as:

\begin{equation}
\label{eq:object camera 2d}
e_{co\_2D} = [\mathbf{c},\mathbf{s}]-[\mathbf{c}_m,\mathbf{s}_m]
\end{equation}

This measurement error has much less uncertainty compared to the 3D error in Eq. \ref{eq:object camera} because 2D object detection is usually more accurate compared to 3D detection. This is similar to projecting map points onto images to formulate reprojection error. However it also loses information after projection because many different 3D cuboids can project to the same 2D rectangle, thus more observations are needed to fully constrain the camera poses and cuboids.

Modelling and estimating the error covariance $\Sigma$ or hessian matrix $W$ is not straightforward compared to points due to the complicated detection process. Therefore we simply give more weights to the semantic confident and geometric close objects. Suppose the cuboid-camera distance is $d$ and the object's 2D detection probability is $p$, then we can define $w=p\times\max(70-d,0)/50$ on KITTI data, where $70m$ is truncation distance. Parameters may vary with different datasets.

\subsubsection{Object-point measurement}
\label{sec:object point}

Points and objects can provide constraints for each other. If point $P$ belongs to an object shown in Fig. \ref{fig:slam measurement}, it should lie inside the 3D cuboid. So we first transform the point to the cuboid frame then compare with cuboid dimensions to get three dimensional error:

\begin{equation}
\label{eq:object point}
e_{op} =   \max(|T_o^{-1}P|-\mathbf{d}_m, \mathbf{0}) 
\end{equation}
\noindent  where $\max$ operator is used because we only encourage points to lie inside cuboid instead of exactly on surfaces.

\subsubsection{Camera-point measurement}
We use the standard 3D point re-projection error in feature-based SLAM \cite{mur2015orb}.
\begin{equation}
\label{eq:point camera}
e_{cp} = \pi(T_c^{-1}P) - z_m
\end{equation}
\noindent where $z_m$ is the observed pixel coordinate of 3D point $P$.

\subsection{Data association}
\label{sec:object association}

\begin{figure}[t]
  \centering
   \includegraphics[scale=0.27]{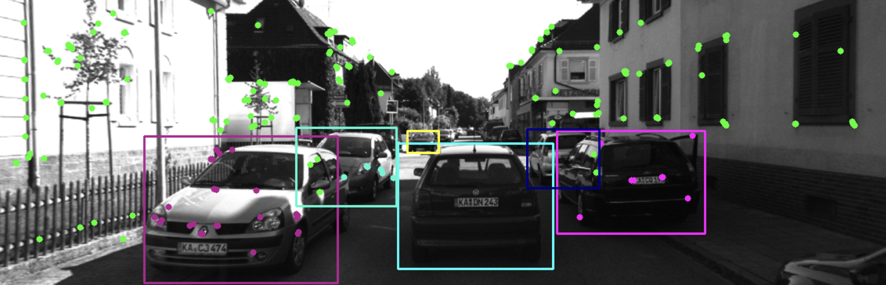}
   \caption{
   Object association in the dynamic and occluded scenarios of KITTI 07. Green points are the non-object points, and points in other colors are associated with objects of the same color. The front cyan moving car is not added as a SLAM landmark as no feature point is associated with it. Points in object overlapping areas are not associated with any object due to ambiguity.
   }
   \label{fig:kitti 07 img}
\end{figure}

Data association across frames is another important part of SLAM. Compared to point matching, object association seems to be easier as more texture is contained and many 2D object tracking or template matching approaches can be used. Even 2D box overlapping can work in some simple scenarios. However, these approaches are not robust if there is severe object occlusion with repeated objects as shown in Fig. \ref{fig:kitti 07 img}. In addition, dynamic objects need to be detected and removed from current SLAM optimization but standard object tracking approaches cannot classify whether it is static or not, unless specific motion segmentation is used.

We thus propose another method for object association based on feature point matching. For many point based SLAM methods\cite{mur2015orb}, feature points in different views can be effectively matched through descriptor matching and epipolar geometry checking. Therefore we first associate feature points to their corresponding object if points are observed in the 2D object bounding box for at least two frames and their 3D distance to the cuboid center is less than 1m. For example in Fig. \ref{fig:kitti 07 img}, the feature points have the same color with their associated object. Note that this object-point association is also used when computing object-point measurement error during BA in Eq. \ref{eq:object point}. Finally, we match two objects in different frames if they have the most number of shared feature points between each other and the number also exceeds certain threshold (10 in our implementation). Through our experiments, this approach works well for wide baseline matching, repetitive objects and occlusions. The dynamic feature points belonging to moving objects are discarded because they cannot fulfill the epipolar constraint. Therefore, objects with few associated feature points are considered as dynamic objects for example the front cyan car in Fig. \ref{fig:kitti 07 img}.


\section{Dynamic SLAM}
\label{sec:dynamic slam}
The previous section deals with the static object SLAM. In this section, we propose an approach to jointly estimate the camera pose and dynamic object trajectories. Our approach makes some assumptions about the objects to reduce the number of unknowns and make the problem solvable. The two commonly used assumptions are that the object is rigid and follows some physically feasible motion model. The rigid body assumption indicates that a point's position on its associated object doesn't change over time. This allows us to utilize the standard 3D map point reprojection error to optimize its position. For the motion model, the simplest form is constant motion model with uniform velocity. For some specific object such as vehicles, it is additionally constrained to follow the nonholonomic wheel model (with some side-slip).



\subsection{Notations}
We define some new map elements in addition to the static SLAM in Sec. \ref{sec:slam pipeline}. For the dynamic object $O^i$, we need to estimate its pose $^j O^i$ in each observed frame $j$. We use ``dynamic points" to refer to feature points associated with moving objects. For dynamic point $P^k$ on moving object $O^i$, we represent its position anchored on the object as $^i P^k$, which is fixed based on the rigidity assumption. Its world pose will change over time and is not suitable for SLAM optimization.

\subsection{SLAM optimization}
\label{sec:dynamic ba}

\begin{figure}[t]
  \centering
   \includegraphics[trim={0cm 11cm 20cm 0cm},clip,scale=0.55]{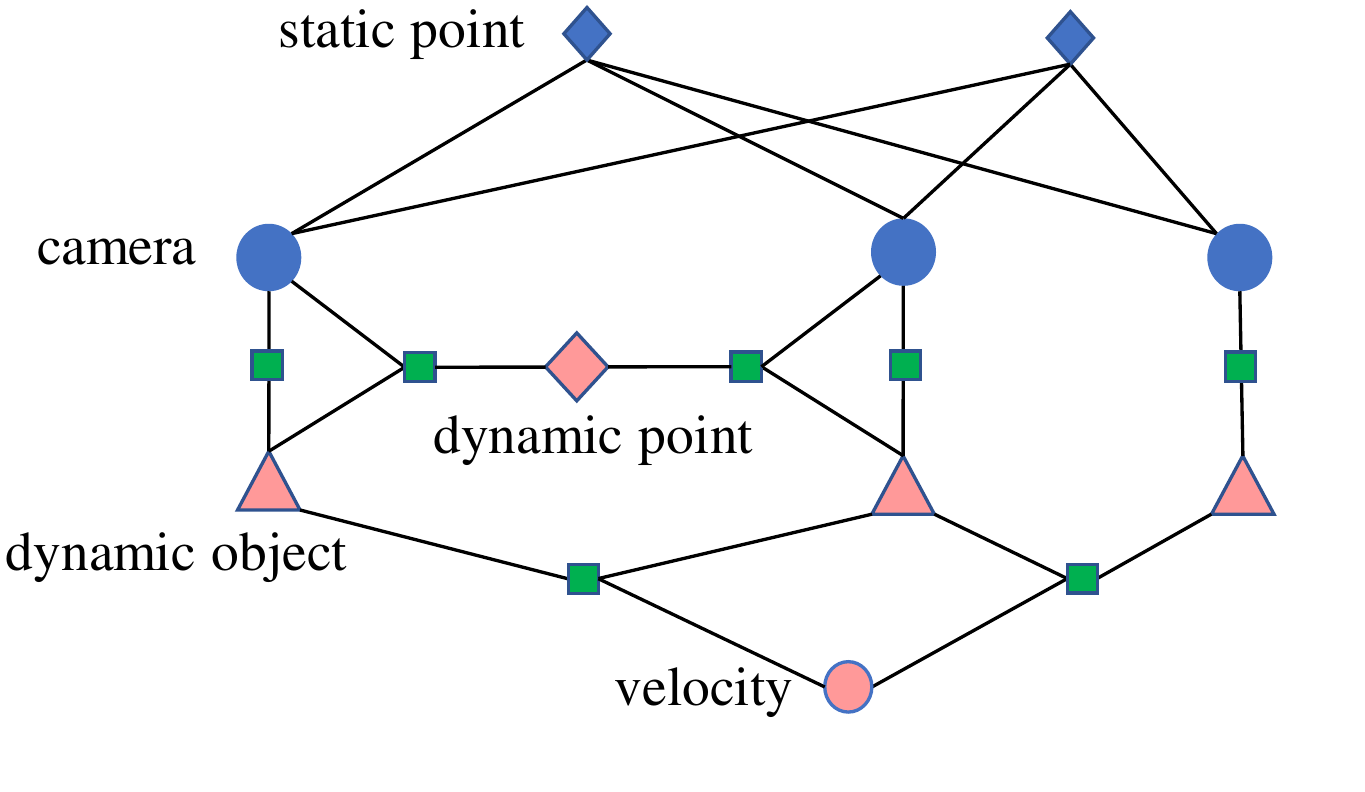}
   \caption{Dynamic object SLAM. Blue nodes represent the static SLAM component and red ones represent new dynamic variables. The green squares are the new factors of dynamic map including motion model constraints and observations of objects and points. }
   \label{fig:dynamic graph}
\end{figure}

The factor graph of the dynamic object estimation is shown in Fig. \ref{fig:dynamic graph}. Blue nodes are the static SLAM components while the red ones represent the dynamic objects, points and motion velocity. The green squares are the measurement factors including the camera-object factor in Eq. \ref{eq:object camera 2d}, the object-velocity factor in Eq. \ref{eq:motion model obs}, and the point-camera-object factor in Eq. \ref{eq:dynamic point obs} which will be explained as follows. With these factors, camera poses can also be constrained by the dynamic elements. 


\subsubsection{Object motion model}
The general 3D object motion can be represented by a pose transformation matrix $T\in SE(3)$. We can apply $T$ to the previous pose then compute pose error with the current pose. Here, we adopt a more restricted nonholonomic wheel model\cite{lavalle2006planning} that is also used in other dynamic vehicle tracking work \cite{li2018stereo}. Car motion is represented by linear velocity $v$ and steering angle $\phi$. Suppose the vehicle moves on a local planar surface approximately, then object roll/pitch$=$0 and translation in z axis $t_z$ = 0. Only $t_x,t_y,\theta$ (heading yaw) is needed to represent its full state $T_o=[R(\theta) \ [t_x, t_y,0]']$. The predicted state from velocity is:

\begin{equation}
\begin{bmatrix}
t_x' \\
t_y' \\
\theta' \\
\end{bmatrix} = 
\begin{bmatrix}
t_x \\
t_y \\
\theta \\
\end{bmatrix} + v\Delta t
\begin{bmatrix}
\cos(\theta) \\
\sin(\theta) \\
\tan(\phi)/L \\
\end{bmatrix}
\end{equation}

\noindent where $L$ is the distance between the front and rear wheel center. Note that this model requires that $x,y,\theta$ is defined at the rear wheel center while our object frame is defined at the vehicle center. The two coordinate frames have $L/2$ offset that needs to be compensated. The final motion model error is simply as:

\begin{equation}
\label{eq:motion model obs}
e_{mo} = [t_x',t_y',\theta']-[t_x,t_y,\theta]
\end{equation}

\subsubsection{Dynamic point observation}

As explained before, the dynamic point is anchored to its associated object, so it is first transformed to the world frame then projected onto the camera. Suppose the local position of $k$th point on $i$th object is $^iP^k$ and the object pose in the $j$th image is $^j T_o^i$, then the point's reprojection error is:

\begin{equation}
\label{eq:dynamic point obs}
e_{dp} = \pi ({^jT_o^i} {^iP^k}, T^j_c) - z_{kj}
\end{equation}

\noindent where $T^j_c$ is the $j$th camera pose and $z_{kj}$ is the observed pixel of this point.

\subsection{Dynamic data association}

Through the experiments, we find that the association method for static environments in Section \ref{sec:object association} is not suitable for the dynamic cases due to the difficulty in matching dynamic point features. The typical way to track a feature point is to predict its projected position, search nearby features match descriptors then check epipolar geometry constraints \cite{mur2015orb}. However, for monocular dynamic cases, it is difficult to accurately predict the movement of objects and points and the epipolar geometry is also not accurate when object motion is inaccurate.

Thus, we designed different approaches for the point and object association. The feature points are directly tracked by the 2D KLT sparse optical flow algorithm, which doesn't require the 3D point position. After pixel tracking, the 3D position of the dynamic features will be triangulated considering the object movement. Mathematically, suppose the projection matrix of two frames are $M_1,M_2$. The 3D point positions in these two frames are $P_1,P_2$ and corresponding pixel observations are $z_1,z_2$. The object movement transformation matrix between two frames is $\Delta T$, then we can infer that $P_2 = \Delta T P_1$. Based on projection rule, we have:

\begin{equation}
\begin{split}
M_1 P_1  = z_1  \\
M_2 \Delta T P_1  = z_2 \\
\end{split}
\end{equation}

\noindent If we treat $M_2 \Delta T$ as a modified camera pose compensating object movement, the above equation is the standard two-view triangulation problem \cite{hartley2003multiple} that can be solved by SVD.

KLT tracking might still fail when the pixel displacement is large, for example when another vehicle comes close and towards the camera. Therefore, for the dynamic object tracking, we do not utilize the shared feature point matching approach in Sec. \ref{sec:object association}. Instead, we directly utilize visual object tracking algorithm \cite{henriques2015high}. The object's 2D bounding box is tracked and its position is predicted from the previous frame, then it is matched to the detected bounding box in the current frame with the largest overlapping ratio.


\section{Implementations} 
\label{sec:implementation}

\subsection{Object detection}
For the 2D object detection, we use the YOLO detector \cite{redmon2016yolo9000} with a probability threshold of 0.25 for indoor scenarios and MS-CNN \cite{cai16mscnn} with a probability of 0.5 for outdoor KITTI. Both run in real time on a GPU.

If an accurate camera pose is known for example in the SUN RGBD dataset, we only need to sample the object yaw to compute the VPs as explained in Section \ref{sec:object sampling}. Fifteen samples of the object yaw in a range of $90 \degree$ are generated as cuboids can be rotated as mentioned in Section \ref{sec:obj cam meas}. Then ten points are sampled on the top edge of the 2D bounding box. Note that not all the samples can form valid cuboid proposals, as some cuboid corners might lie outside of the 2D box. In scenarios with no ground truth camera pose provided, we sample camera roll/pitch in a range of $\pm 20\degree$ around the initially estimated angle. For single images with no prior information, we simply estimate that camera is parallel to ground. For multi-view scenarios, SLAM is used to estimate the camera pose. One advantage of our approach is that it doesn't require large training data as we only need to tune the two cost weights in Eq. \ref{eq:score proposal}. It can also run in real time, including the 2D object detection and edge detection.


\subsection{Object SLAM}
\label{sec:SLAM implementation}

The pipeline of the whole SLAM algorithm is shown in Fig. \ref{fig:slam flow}. As mentioned in Sec. \ref{sec:object slam}, our system is based on ORB SLAM2 and we didn't change the camera tracking and keyframe creation modules. For the newly created keyframe, we detect the cuboid objects, associate them, then perform bundle adjustment with camera poses and points. For the dynamic objects, we can choose to reconstruct or ignore them depending on different tasks. The cuboid is also used to initialize the depth for feature points that are difficult to triangulate when stereo baseline or parallax angle is smaller than a threshold. This can improve the robustness in some challenging scenarios such as large camera rotations as demonstrated in the experiments. Since the number of objects is far less than points, object association and BA optimization runs efficiently in real time. To get an absolute map scale for monocular SLAM, the initial frame's camera height is provided to scale the map. Note that our object SLAM can also work independently without points. In some demonstrated challenging environments with few feature points, ORB SLAM cannot work, but our algorithm can still estimate camera poses using only the object-camera measurement.


There are different costs in the optimization (see in Sec. \ref{sec:object slam}) and some of them are in pixel space for example Eq. \ref{eq:object camera 2d} while some are in Euclidean space such as Eq. \ref{eq:object camera} and \ref{eq:object point}, therefore it is necessary to tune the weights between them. Since it is difficult to analyze the cuboid detection uncertainty, we mainly hand-tune the object cost weights by inspecting the number and magnitude of measurement so that different types of measurements contribute roughly the same. For example, there are only a few objects compared to points but their reprojection error in Eq. \ref{eq:object camera 2d} is much larger compared to points. From our experiments, object-camera and point-camera measurements have similar weights.

\subsection{Dynamic object}
\label{sec: dynamic implementation}

The implementation of dynamic objects mostly follows the previous section with some difference. The constant motion model assumption may not hold for practical datasets because objects may accelerate and decelerate (for example in Fig. \ref{fig:dynamic velocity esti}). Through the ground truth object velocity analysis, we find that the velocity roughly stays the same in about 5 seconds. Therefore, in our SLAM, motion model constraint is only applied to observations in the last 5 seconds.

\section{Experiments - Single View Detection}
\label{sec:experiment single view}

The SUN RGBD \cite{song2015sun} and KITTI object \cite{geiger2012we} data with ground truth 3D bounding box annotations are used for single view object detection evaluation. 3D intersection over union (IoU) and average precision (AP) is adopted as the evaluation metric instead of only rotation or viewpoint evaluation in many other works. As there is no depth data, the 3D IoU threshold for a correct detection is adjusted to 25\% \cite{song2015sun,chen2016monocular}. Since our approach doesn't depend on the prior object model, in order to get an absolute scale of object position and dimensions, we only evaluate the ground objects with known camera height as explained in Sec. \ref{sec:proposal generation}. For the KITTI dataset, this assumption is already satisfied. The commonly used training and validation index split by \cite{xiang2017subcategory,mousavian20163d} is used. For the SUN RGBD dataset, we select 1670 images with a visible ground plane and ground objects fully in the field of view.

\subsection{Proposal Recall}
\begin{figure}[t]
  \centering
  \subfigure[]{\includegraphics[scale=0.55]{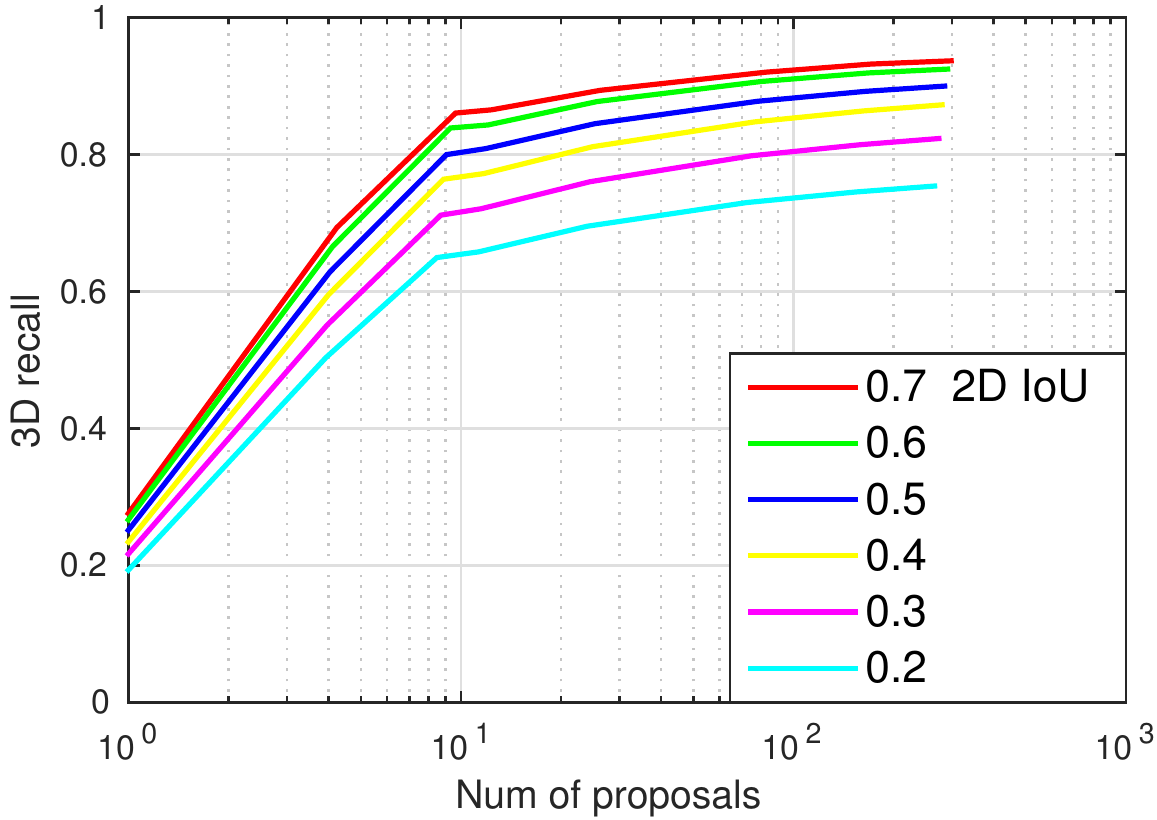} \label{fig:sun recall} }
  \subfigure[]{\includegraphics[scale=0.60]{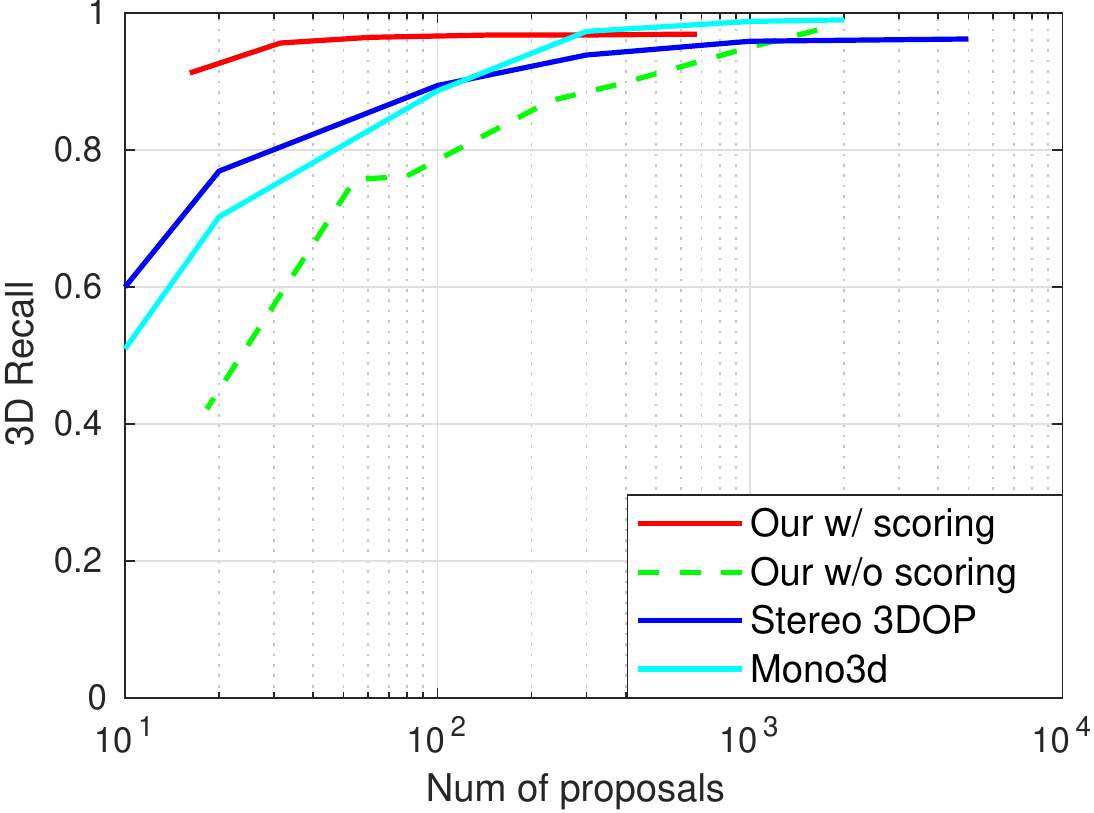} \label{fig:kitti recall}}
   \caption{(a) 3D proposal recall on SUN RGBD Subset dataset. Different lines correspond to different 2D box IoU for association. (b) 3D proposal recall on KITTI dataset. Our approach can get higher recall with fewer proposals.}
   \label{fig:sun kitti recall}
\end{figure}

We first evaluate the quality of proposal generation in SUN RGBD. It is obvious that if the 2D bounding box is inaccurate, our 3D cuboid accuracy will also be affected. This effect is analyzed by evaluating the 3D recall on objects with a 2D IoU greater than a threshold $\tau$, as shown in Fig. \ref{fig:sun recall}. As expected, a larger $\tau$ leads to a higher 3D recall. Our approach can achieve 3D recall of 90\% using around 50 cuboid proposals when 2D IoU is $0.6$.


We then evaluate and compare the proposal quality on the KITTI dataset shown in Fig. \ref{fig:kitti recall}. Since Mono3d \cite{chen2016monocular} and 3DoP \cite{chen20153d} utilize different validation indexes compared to us, we only evaluate on the common images (1848). From our tests, different image indexes only lead to small result changes. Results of other algorithms are taken from their paper. Note that Mono3d first exhaustively samples huge amounts of cuboid proposals ($\sim$14k), and then reports the recall after scoring and selecting the top proposals based on semantic and instance segmentation. Therefore, we also evaluate the recall before and after scoring. Before scoring (green line), our approach can reach a recall of 90\% with 800 raw proposals per image, about 200 proposals per object. After scoring (red line), we can reach the same recall using just 20 proposals, much fewer compared to \cite{chen2016monocular}. There are two main reasons for this. First, our 3D proposals are of high quality because they are guaranteed to match the 2D detected box. Second, our more effective scoring function. Note our approach has an upper limit as shown in Fig. \ref{fig:kitti recall} because the 2D detector might miss some objects.



\begin{table}[t]
\caption{Comparison of 3D Object Detection on SUN RGBD Subset and KITTI Dataset}
\begin{center}
\begin{threeparttable}
\begin{tabular}{c | c | c @{\hskip 1em} c c}
\hline
& Method            & 3D IoU	& AP		  \\   \hline
&Primitive\cite{xiao2012localizing}  & 0.36	 				 	& 0.27				        \\
SUN &3dgp\cite{choi2013understanding}& 0.42	 	 	 			& 0.22	            		\\
RGBD &Ours							 & 0.39  		    	 	& 0.27			      	  	\\
&Ours\tnote{*}						 & 0.45						& 0.30			      	  	\\ \hline
\multirow{5}{*}{KITTI} &Deep3D\cite{mousavian20163d} 	& 0.33	 	& 0.69		   		    \\ 
&SubCNN\cite{xiang2017subcategory}	 & 0.21	 	 	 			& 0.17		            	\\ 
&Mono3D\cite{chen2016monocular}      & 0.22						& 0.27							\\
&Ours								 & 0.21  		    		& 0.29 		 	      	  \\
&Ours top 10						 & 0.38	      	 		    & 0.75	      	  	\\
\hline
\end{tabular}
\begin{tablenotes}
\footnotesize
\item[*] On 3dgp detected images.
\end{tablenotes}
\end{threeparttable}       
\end{center}
\label{table:rgbd kitti error}
\end{table}

\begin{figure}[t]
  \centering
   \includegraphics[scale=0.455]{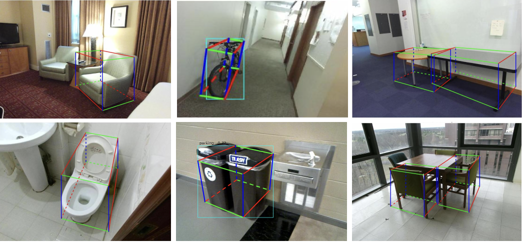}\\
    \vspace{0.5em}
   \includegraphics[scale=0.38]{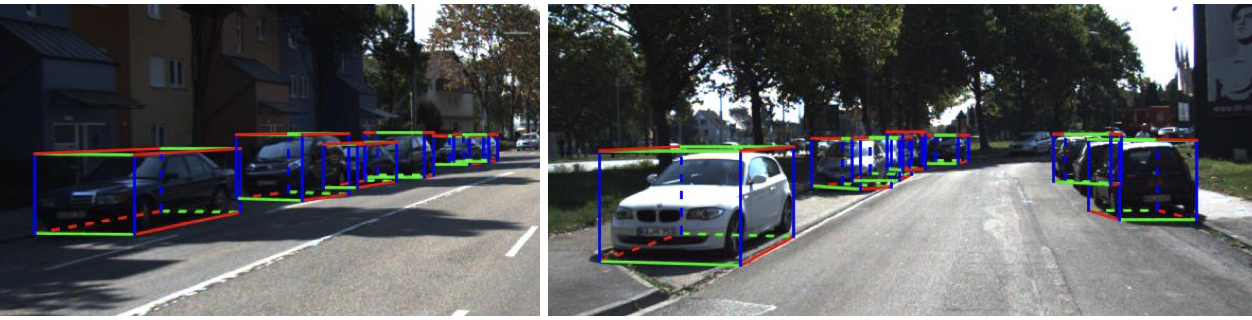}
   \caption{Single image 3D object detection examples in SUN RGBD and KITTI dataset.}
   \label{fig:single image examples}
\end{figure}

\subsection{Final detection}
We then evaluate the final accuracy of the best selected proposal. To the best of our knowledge, there is no trained 3D detection algorithm in SUN RGBD. Therefore, we compared with two public approaches, SUN primitive \cite{xiao2012localizing} and 3D Geometric Phrases (3dgp) \cite{choi2013understanding}. Both of which are model based algorithms like ours. Additionally, 3dgp uses fixed prior object models. We modify their code to use the actual camera pose and calibration matrix when detecting and unprojecting to 3D space. 

For 3D IoU evaluation, in order to eliminate the effect of 2D detector, we only evaluate 3D IoU for objects with 2D rectangle IoU$>0.7$. As shown in Table \ref{table:rgbd kitti error} and Fig. \ref{fig:single image examples}, our approach can generate many more accurate cuboids. Other approaches \cite{xiao2012localizing,choi2013understanding} can only detect around 200 cuboids in the SUN RGBD subset datasets while our algorithm detects ten times more. Our mean 3D object IoU is smaller compared 3dgp using prior models, but higher if we only evaluate on the same detected images ($\approx$200) by 3dgp. Similarly for average precision, we evaluate other methods only on the images where they detect 3D objects, otherwise their AP value will be very low ($<$5\%) compared to our 27\% on the whole dataset.


On the KITTI dataset, we compare with other monocular algorithms \cite{mousavian20163d,xiang2017subcategory,chen2016monocular} using deep networks. SubCNN additionally uses prior models. Prediction results are provided by their authors. AP is evaluated on the moderate car class. From Table \ref{table:rgbd kitti error}, our approach performs similarly to SubCNN and Mono3d. As SubCNN generates many false positive detections therefore their AP value is low. The best performing approach is Deep3D \cite{mousavian20163d} which directly predicts vehicle orientations and dimension using deep networks. As there is only one object class ``car" with fixed camera poses and object shapes, CNN prediction works better than our hand-designed features. The last row is the evaluation of our selected top ten cuboid proposals to show that our proposal generation part can still generate high quality proposals.

\section{Experiments - Object SLAM}
\label{sec:experiment object SLAM}

We then evaluate the performance of object SLAM, including camera pose estimation, and 3D object IoU after BA optimization. We show that SLAM and object detection can benefit each other in various datasets. Root mean squared error (RMSE) \cite{sturm2012benchmark} and KITTI translation error \cite{geiger2012we} are used to evaluate the camera pose. Note, even though our algorithm is monocular SLAM, we can get the map scale from the first frame's camera height, therefore, we directly evaluate the absolute trajectory error without aligning it in scale. To better evaluate the monocular pose drift, we turn off the loop closure module in ORB SLAM when using and comparing with it.


\subsection{TUM RGBD and ICL-NUIM dataset}
These datasets \cite{sturm2012benchmark,handa:etal:ICRA2014} have ground truth camera pose trajectory for evaluation. We only use the RGB images for the SLAM algorithm. For ground truth of the objects, 3D cuboids are manually labeled in a registered global point cloud from the depth images.

\begin{figure}[t]
  \centering
   \includegraphics[scale=0.20]{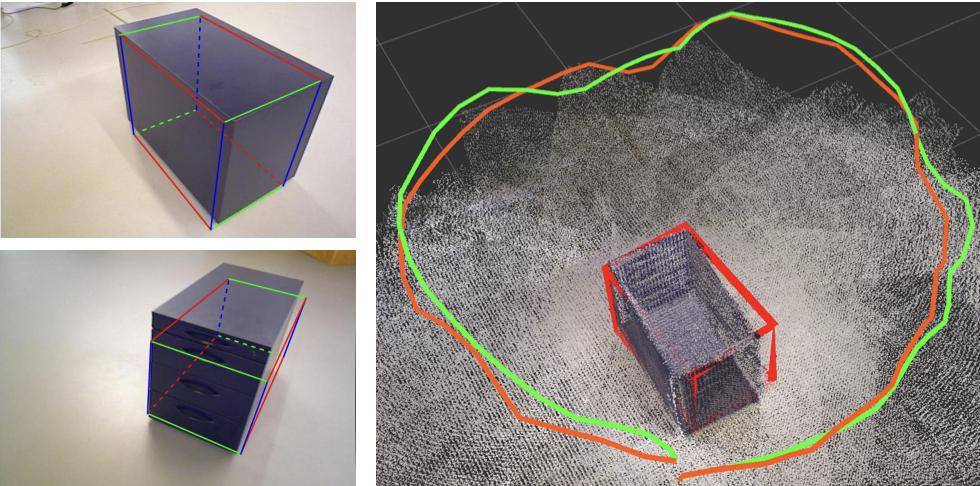} 
   \caption{Object SLAM on TUM fr3\_cabinet. Red cube is the optimized object landmark, matching well with the ground truth point cloud. Red and green trajectories are the predicted and ground truth camera paths. Existing SLAM algorithms fail on this dataset due to low texture. Only object is used in SLAM.}
   \label{fig:tum object slam}
\end{figure}

We first test on TUM \textit{fr3\_cabinet}, shown in Fig. \ref{fig:tum object slam} which is a challenging low texture dataset, and existing monocular SLAM algorithms all fail on it due to few point features. The object is the only SLAM landmark and the 3D object-camera measurement in Sec. \ref{sec:obj cam meas} is used because it can provide more constraints than 2D measurement. The left of Fig. \ref{fig:tum object slam} shows our online detected cuboid in some frames using estimated camera pose from SLAM. There is clearly large detection error in the bottom image. After multi-view optimization, the red cube in the map almost matches with the ground truth point cloud. From row ``fr3/cabinet" in Table. \ref{table:tum chair error}, 3D object IoU is improved from 0.46 to 0.64 after SLAM optimization compared to the single image cuboid detection. The absolute camera pose error is 0.17m. 

We then test on the ICL living room dataset which is a general feature rich scenario. Since there is no absolute scale for monocular DSO or ORB SLAM, we compute their pose error after scale alignment \cite{engel2017direct}. We improve the object detection accuracy while sacrificing some camera pose accuracy due to imperfect object measurements. As can be seen from the mapping result of ICL data in Fig. \ref{fig:icl mapping}, our approach is able to detect different objects including sofas, chairs, and pot-plant demonstrating the advantage of our 3D detection without prior models.


\begin{table}
\caption{Object Detection and SLAM Result on Indoor Datasets}
\begin{center}
\begin{threeparttable}
\begin{tabular}{c | c c | c c c}
\hline
\multirow{2}{*}{Dataset}   &\multicolumn{2}{c|}{Object IoU}   &\multicolumn{3}{c}{Pose error (m)} \\ 
			& single view   & after BA		 	  & DSO \tnote{*} 		& ORB \tnote{*}   & Our \\ \hline 
fr3 cabinet 		& 0.46	 	& \textbf{0.64}      & ---   			& ---    & \textbf{0.17}       \\ 
ICL room2 		& 0.33	 	& \textbf{0.49}      & \textbf{0.01}   		& 0.02  & 0.03        \\ 
Two Chair 		& 0.37	 	& \textbf{0.58}       & 0.01   		& --- & \textbf{0.01}       \\ 
Rot Chair 		& 0.35	 	& \textbf{0.50}       & ---   		& ---    & \textbf{0.05}       \\ 
\hline
\end{tabular}
\begin{tablenotes}
\footnotesize
\item[*] Pose error with scale alignment.
\end{tablenotes}
\end{threeparttable}       
\end{center}
\label{table:tum chair error}
\end{table}

\subsection{Collected chair dataset}
\begin{figure}[t]
  \centering
   \subfigure[]{\includegraphics[scale=0.15]{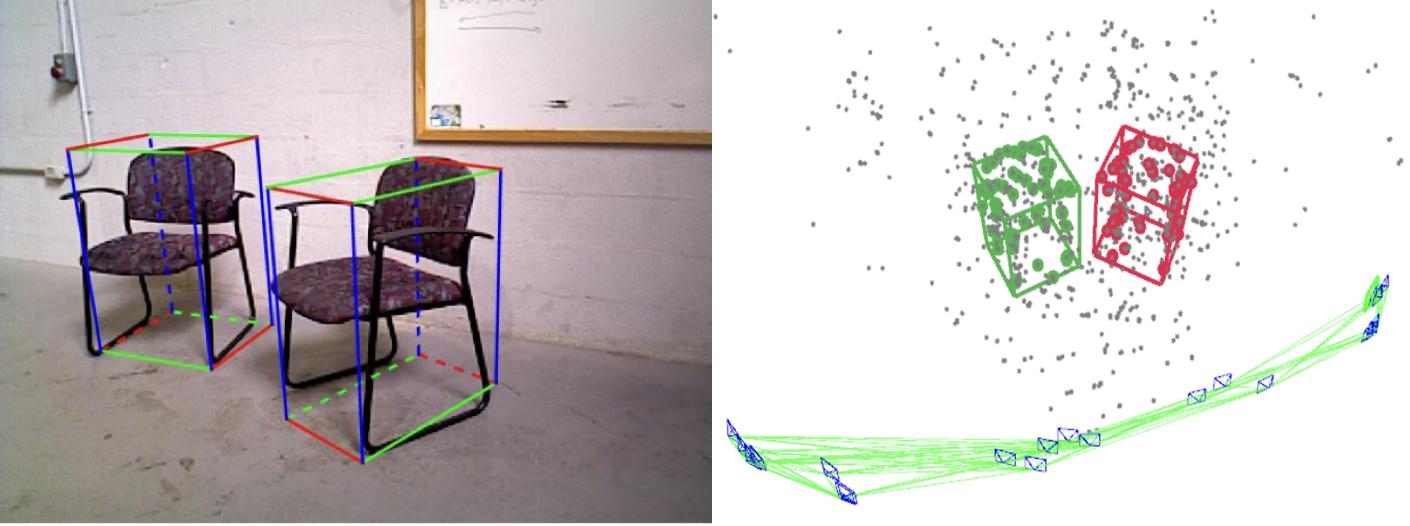} \label{fig:two chair}}
   \subfigure[]{\includegraphics[scale=0.125]{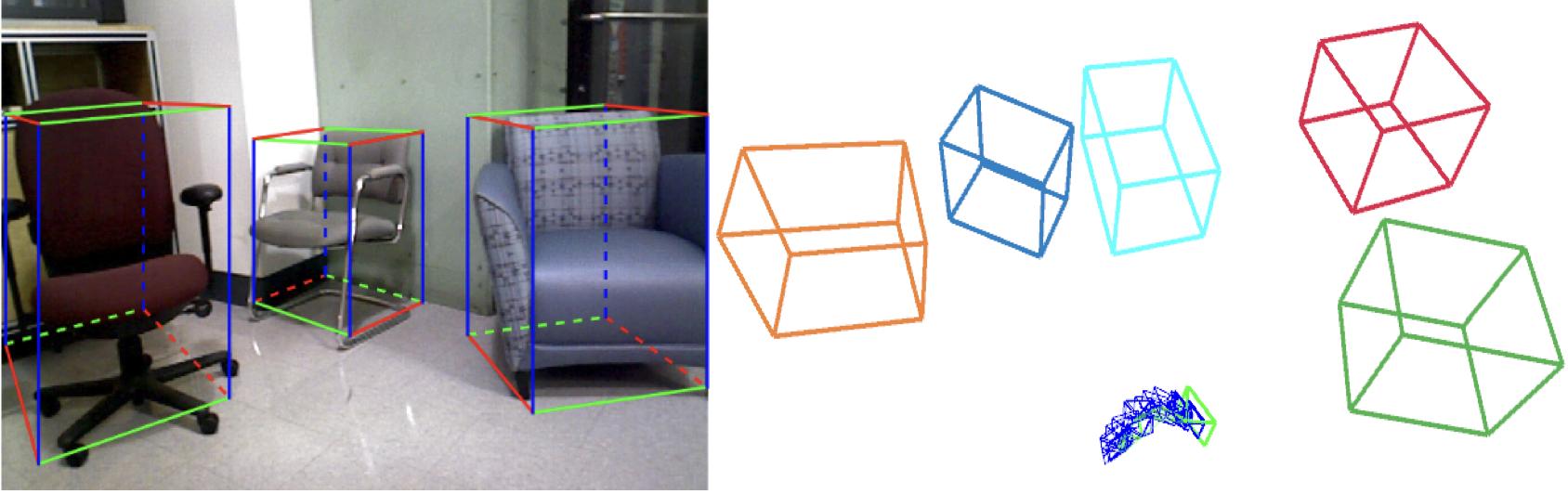} \label{fig:five chair}}
   \caption{ Collected chair datasets. (a) Objects fit tightly with the associated points after BA optimization. (b) Objects can improve camera pose estimation when there is large camera rotation.   
   }
   \label{fig:collected chair}   
\end{figure}

We collect two chair datasets using a Kinect RGBD camera shown in Fig. \ref{fig:collected chair}. The RGBD ORB SLAM result is used as the ground truth camera poses. The second dataset contains large camera rotation which is challenging for most monocular SLAM.  As shown in Fig. \ref{fig:two chair}, after optimization, cuboids can fit the associated 3D points tightly showing that object and point estimation benefit each other. The quantitative error is shown in the bottom two rows of Table. \ref{table:tum chair error}. DSO is able to work in the first dataset but performs poorly in the second one, due to the large camera rotation. Mono ORB SLAM fails to initialize in both cases while our cuboid detection can provide depth initialization for points even from a single image. Similar as before, the 3D object IoU is also improved after BA.

\begin{table*}[t]
\caption[Object detection and camera pose estimation on KITTI raw sequence]{Object Detection and Camera Pose Estimation on KITTI Raw Sequence}
\begin{center}
\begin{tabular}{c | c | c c c c c c c c c c | c}
\hline
\multicolumn{2}{c|}{Seq}   &22  &23  &36  &39  &61  &64  &93  &95  &96  &117  &Mean \\ \hline
\multirow{3}{*}{\shortstack{Object\\ 3D IoU}}  &Single view\cite{mousavian20163d}  &0.52  &0.32   &0.50  &0.54   &\textbf{0.54}   &0.43   &0.43   &0.40   &0.26   &0.25  &0.42\\
&Ours before BA &0.55 &\textbf{0.36} &0.49 &0.56 &0.54 &0.42 &\textbf{0.46}  &0.49  &0.20 &0.30 &0.44 \\
   &Ours 	  &\textbf{0.58}   &0.35   &\textbf{0.54}  & \textbf{0.59}   &0.50   &\textbf{0.48}   &0.45   &\textbf{0.52}   &\textbf{0.29}   &\textbf{0.35}  &\textbf{0.47} \\ \hline
Trans  &ORB -No LC   &13.0   &\textbf{1.17}   &7.08  &6.76   &\textbf{1.06}   &7.07   &4.40  &\textbf{0.86}  &3.96  &4.10 &4.95\\
error(\%)    &Ours 	 &\textbf{1.68}   &1.72  &\textbf{2.93}   &\textbf{1.61}   &1.24   &\textbf{0.93}   &\textbf{0.60}  &1.49   &\textbf{1.81}   &\textbf{2.21}  &\textbf{1.62}\\ \hline
\end{tabular}
\end{center}
\label{table:kitti raw error}
\end{table*}

\setlength\doublerulesep{0.6em}
\begin{table*}[t]
\caption{Camera Pose Estimation Error on KITTI Odometry Benchmark}
\begin{center}
\begin{tabular}{c| c | c | c c c c c c c c c c | c}
\hline
\multicolumn{3}{c|}{Seq}   &0  &2  &3  &4  &5  &6  &7  &8  &9  &10 &Mean \\ \hline
\multirow{6}{*}{\shortstack{Trans\\ Error \\ (\%)}} &\multirow{3}{*}{\shortstack{Ground\\ based}}  &\cite{lee2015online} 	  & 4.42  &4.77   &8.49   &6.21   &5.44   &6.51   &6.23    &8.23   &9.08   &9.11 & 6.86\\
 &   &\cite{song2016high} 	  &\textbf{2.04}   &\textbf{1.50}   &3.37   &1.43   &2.19   &\textbf{2.09}   & ---      &\textbf{2.37}   &\textbf{1.76}   &\textbf{2.12} & 2.03 \\
&   &Ours 	  &\textbf{1.83}   &\textbf{2.11}   &2.55   &1.68   &2.45   &\textbf{6.31}   & 5.88      &\textbf{4.2}   &\textbf{3.37}   &\textbf{3.48} & 3.39 \\ \cline{2-14}
 &Object  &\cite{sucar2017bayesian}  &3.09   &6.18   &3.39   &32.9  &4.47   &12.5  &2.81    &4.11   &11.2  &16.8  & 9.75\\  
 &based    &Ours 				  &2.40   &4.25   &\textbf{2.87}   &\textbf{1.12}   &\textbf{1.64}   &3.20   &\textbf{1.63}  &2.79   &3.16   &4.34  & 2.74\\ \cline{2-14}
&Combined  &Ours 	  &1.97  &2.48   &1.62   &1.12    &1.64   &2.26   &1.63    &2.05    &1.66    &1.46  &\textbf{1.78} \\ \hline 
\hline

RMSE &Object   &\cite{frost2018recovering} 	  &73.4  &55.5  &30.6  &10.7  &50.8  &73.1  &47.1  &72.2  &31.2  &53.5  &49.8\\
(m) &based  &Ours  &13.9  &26.2  &3.79  &1.10  &4.75  &6.98  &2.67  &10.7  &10.7  &8.37  &\textbf{8.91} \\
\hline

\end{tabular}
\end{center}
\label{table:kitti odom error}
\end{table*}

\subsection{KITTI Dataset}
\begin{figure*}[t]
  \centering
  \subfigure[Sequence 00]{\includegraphics[width=0.26\textwidth,height=0.22\textwidth]{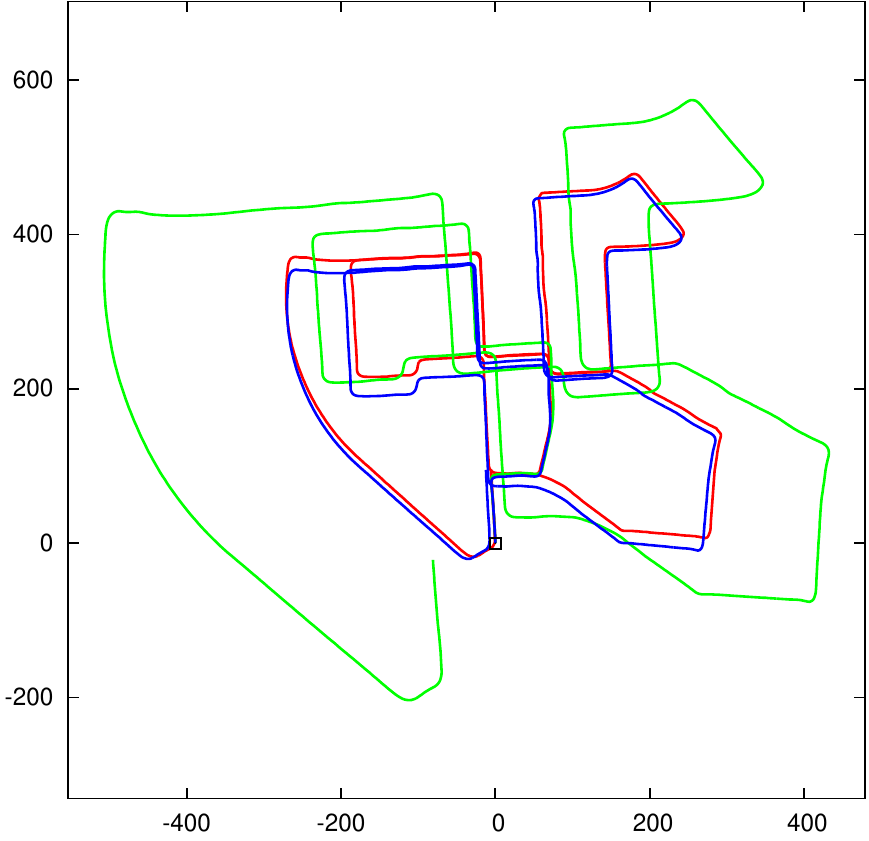}}
  \subfigure[05]{\includegraphics[width=0.26\textwidth,height=0.22\textwidth]{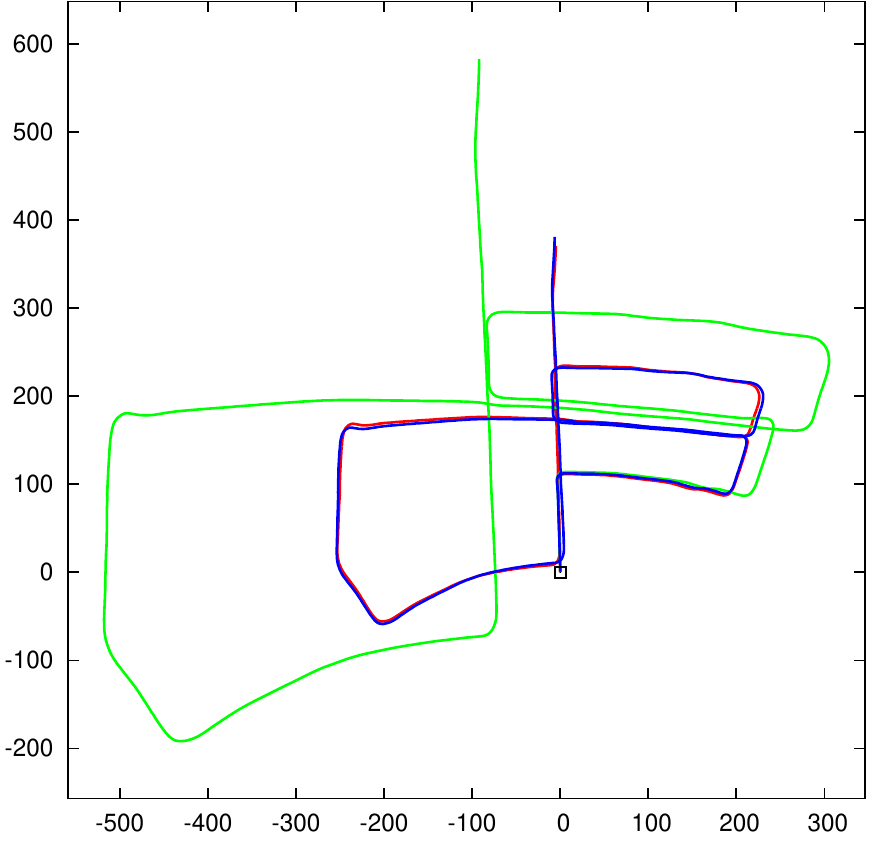}}
  \subfigure[07]{\includegraphics[width=0.26\textwidth,height=0.22\textwidth]{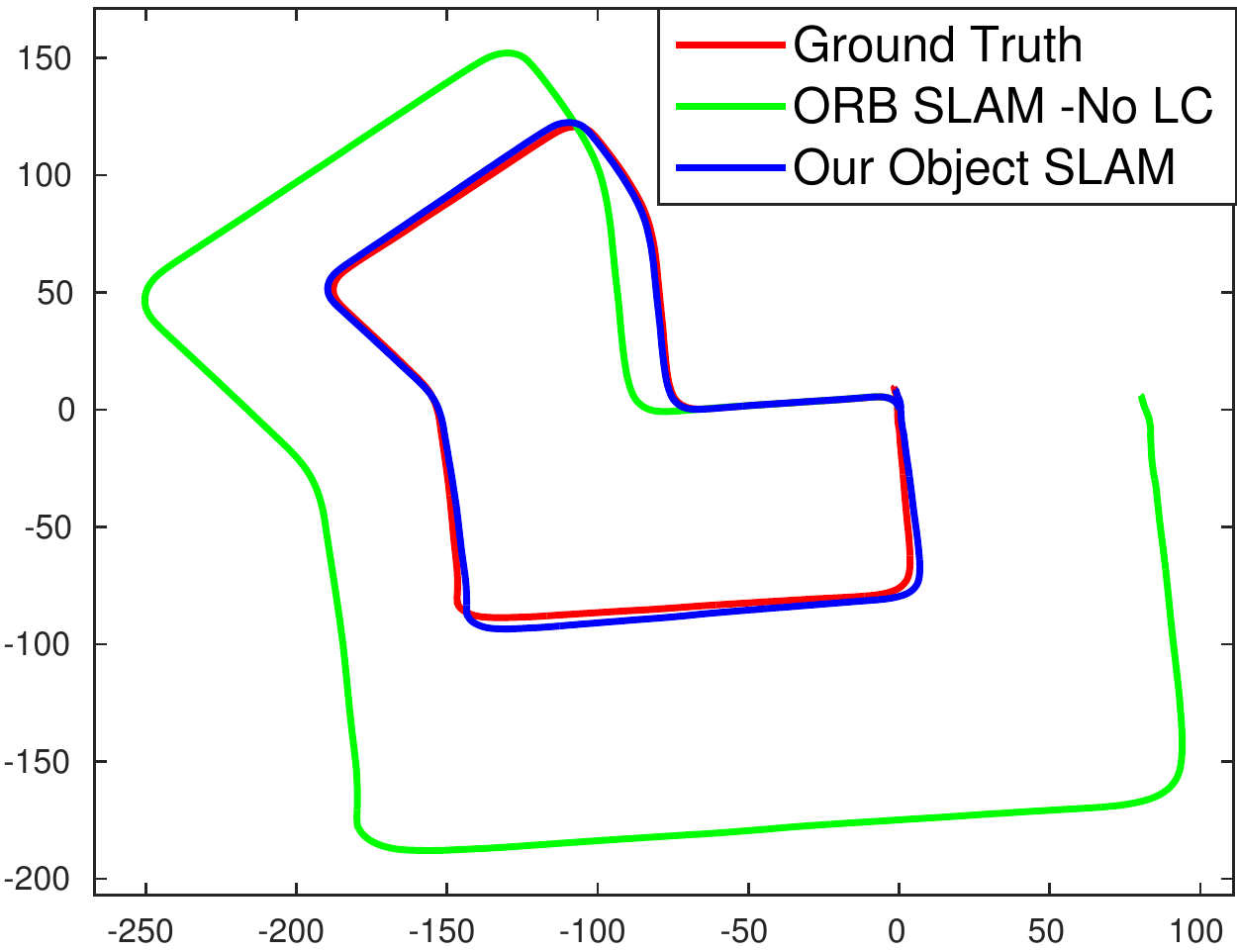}}\vspace{-0.5em}
  \subfigure[06]{\includegraphics[width=0.22\textwidth,height=0.20\textwidth]{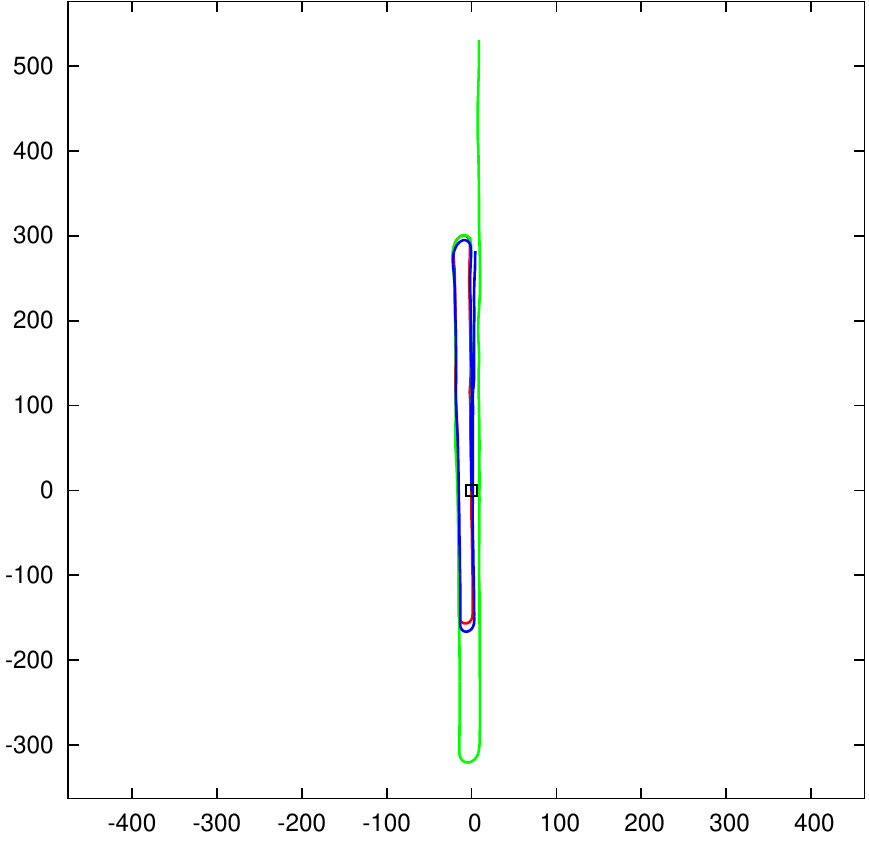}}
  \subfigure[08]{\includegraphics[width=0.22\textwidth,height=0.20\textwidth]{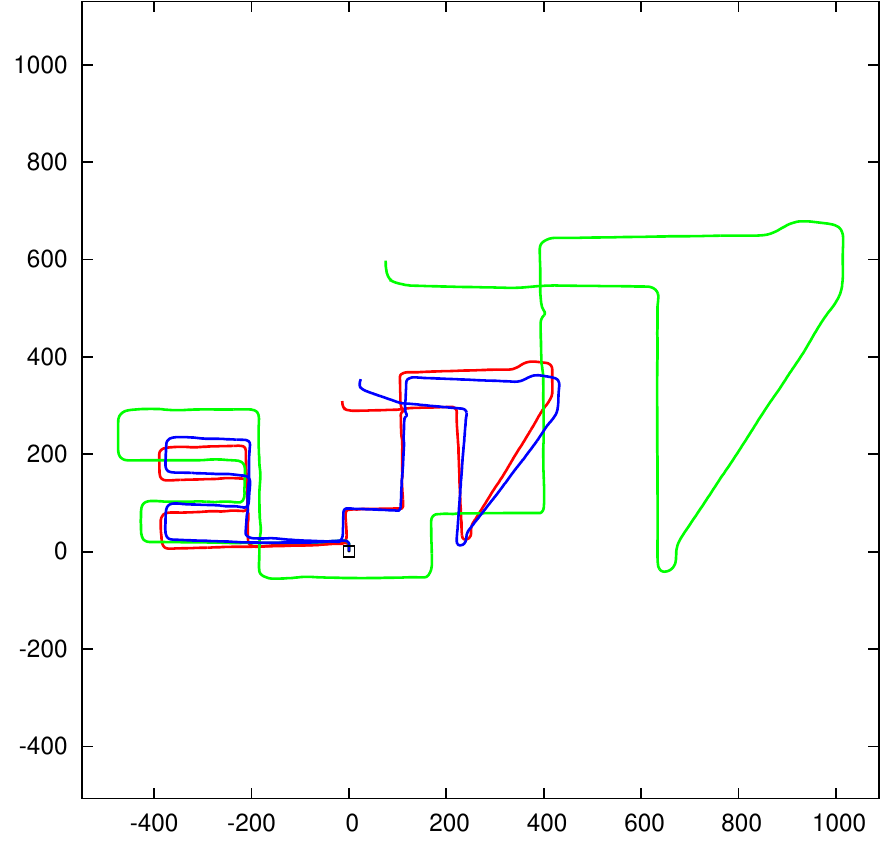}}
  \subfigure[09]{\includegraphics[width=0.22\textwidth,height=0.20\textwidth]{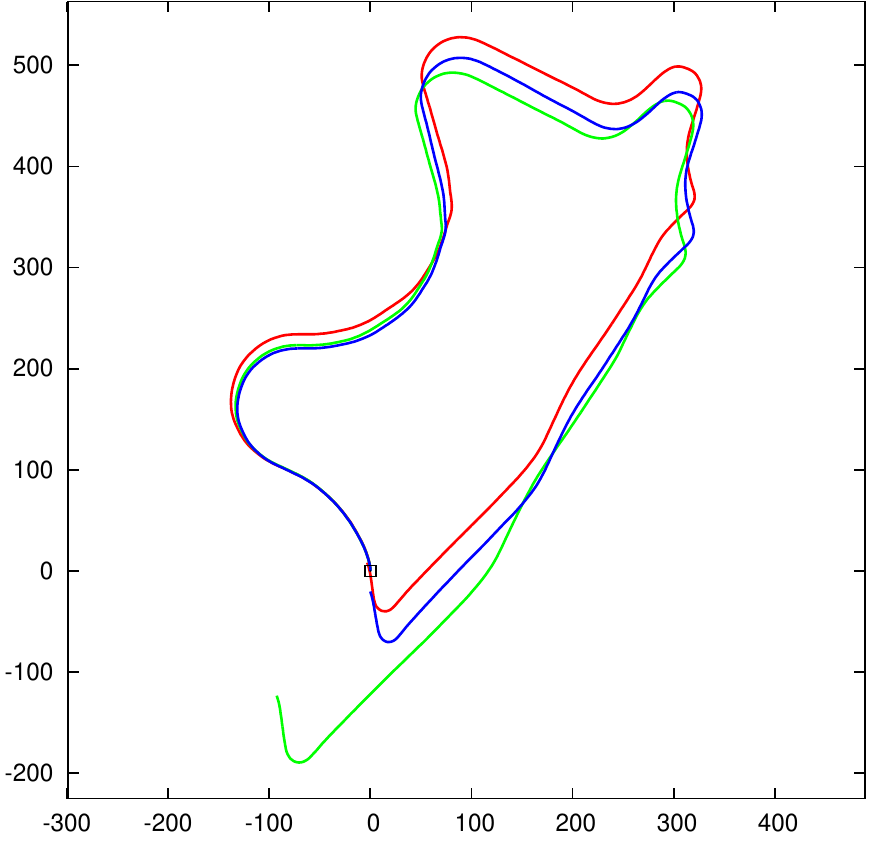}}
  \subfigure[10]{\includegraphics[width=0.22\textwidth,height=0.20\textwidth]{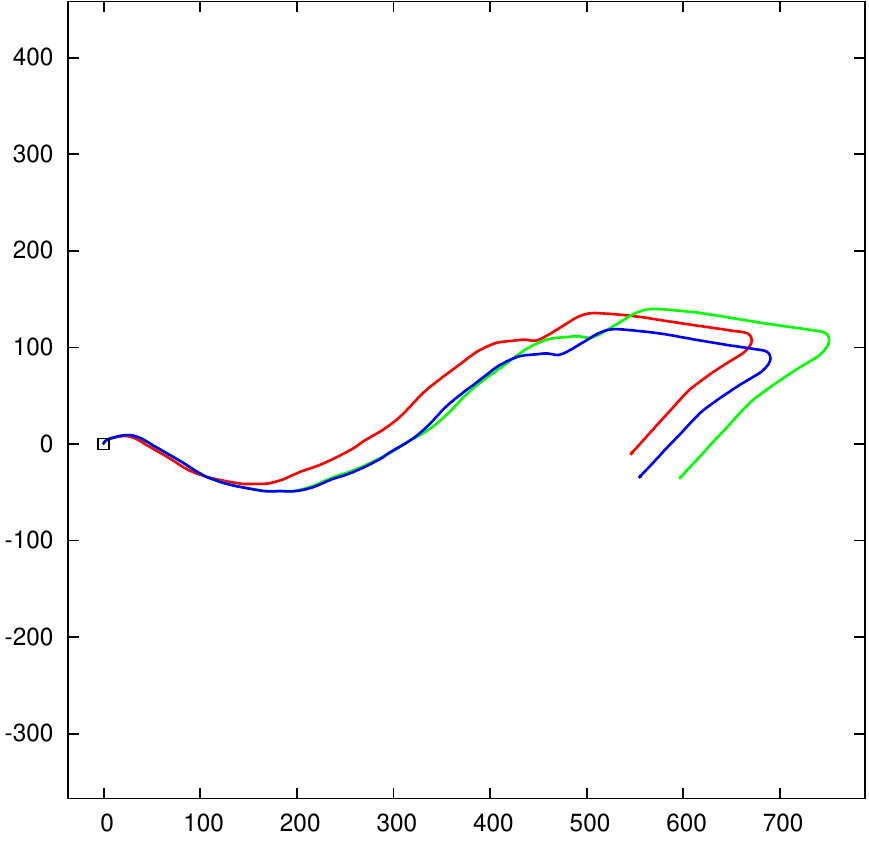}}
   \caption{Our object SLAM on KITTI odometry benchmark without loop closure detection and constant ground height assumption. Red is ground truth pose estimation. Blue is our object SLAM result. Green is ORB SLAM without loop detection. Objects can reduce monocular scale drift and improve pose estimation accuracy.}
  \label{fig:kitti result path}
\end{figure*}

We tested on two of the KITTI datasets, the short sequence, with provided ground truth object annotations, and the long sequence, which is a standard odometry benchmark without object annotations. The 2D object-camera measurement in Sec. \ref{sec:obj cam meas} is used for BA because of its low uncertainty compared to 3D measurements for vehicle detection. We also scale ORB SLAM's initial map by the first frame camera height (1.7m in our implementations) in order to evaluate its absolute pose error. In Fig. \ref{fig:kitti result path}, we can observe that the initial trajectory segment before first turning matches well with ground truth, indicating the initial map scaling for ORB is correct. For KITTI dataset, we additionally initialize object dimension using prior car size ($w=3.9,l=1.6,h=1.5$ in our implementation) to maintain long-term scale consistency, which is also used in other object SLAM works \cite{frost2018recovering,sucar2017bayesian}. This is especially useful when objects are not observed frequently in some sequence.

\subsubsection{KITTI raw sequence} We select 10 KITTI raw sequences with the most number of ground truth object annotations named ``2011\_0926\_00xx''. The ground truth camera pose is from the provided GPS/INS poses on KITTI. For the object IoU, we compare three methods. The first is the single image cuboid detection \cite{mousavian20163d}. The second is the object pose just using SLAM data association between frames, shown as row ``Ours before BA". For example, if an object in one frame is far away, the 3D detection may be inaccurate but in another frame, the same object is closer thus the 3D detection becomes more accurate. Therefore, data association with correct camera pose estimation should also improve 3D detection. Thirdly, the object poses after our final BA optimization are also evaluated shown as row ``Ours".

As shown in the top three rows of Table \ref{table:kitti raw error}, object accuracy is increased after data association and BA optimization in most of the sequences, however, in some sequences, due to local position drift, the object IoU may also decrease a bit. For camera pose estimation, object SLAM can provide geometry constraints to reduce the scale drift of monocular SLAM. Note that since most KITTI raw sequences don't have loops, disabling or enabling the ORB SLAM loop closure module does not make a difference.

\subsubsection{KITTI odometry benchmark}
Most existing monocular SLAMs use a constant ground plane height assumption on the benchmark to reduce monocular scale drift \cite{lee2015online,song2016high}. Recently, there are also some object based scale recovery approaches \cite{frost2018recovering,sucar2017bayesian}. Their results are directly taken from their papers. Similar to other approaches, we didn't compare with ORB SLAM in this case, as without loop closure, it cannot recover scale in the long sequence and has significant drift error shown in Fig. \ref{fig:kitti result path}. As shown in Table \ref{table:kitti odom error}, our object SLAM achieves 2.74 \% translation error and performs much better than other SLAM using objects. This is because they represent vehicles as spheres or only use vehicle height information, which is not as accurate as our cuboid BA. Our algorithm is also comparable to ground-based scaling approaches. Visualization of some object mapping and pose estimation are shown in Fig. \ref{fig:kitti mapping} and Fig. \ref{fig:kitti result path}, where we can see that our approach greatly reduces monocular scale drift. 

Our object SLAM performs worse in some sequences such as Seq 02, 06, 10, mainly because there are not many objects visible over long distance causing large scale drift. Therefore, we also propose a simple method to combine ground height assumption with our object SLAM. If there are no objects visible in recent 20 frames, we fit a ground plane from point cloud then scale camera poses and local map based the constant ground height assumption. As shown in Table \ref{table:kitti odom error}, the fourth row ``ground based/Ours" representing ORB SLAM only with our simple ground scaling and without objects, achieves good performance with 3.39\% translation error. The row ``Combined", which combines ground scaling with objects, further reduces the error to 1.78\% and achieves the state-of-art accuracy of monocular SLAM on the KITTI benchmark. Note that ground plane based approaches also have their limitations for example they won't work for aerial vehicle or handheld cameras. They will also fail when the ground is not visible such as frames in Fig. \ref{fig:kitti 07 img} of KITTI 07. The front dynamic car occludes the ground for a long time and therefore many ground-based approaches fail or perform poorly on KITTI 07.

\begin{figure}[t]
  \centering
   \subfigure[]{ \includegraphics[scale=0.27]{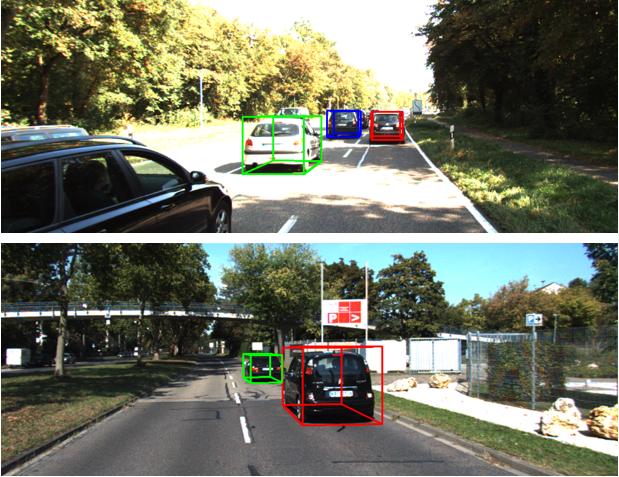}   \label{fig:dynamic sample image}}
   \subfigure[]{ \includegraphics[width=1.9cm,height=4.5cm]{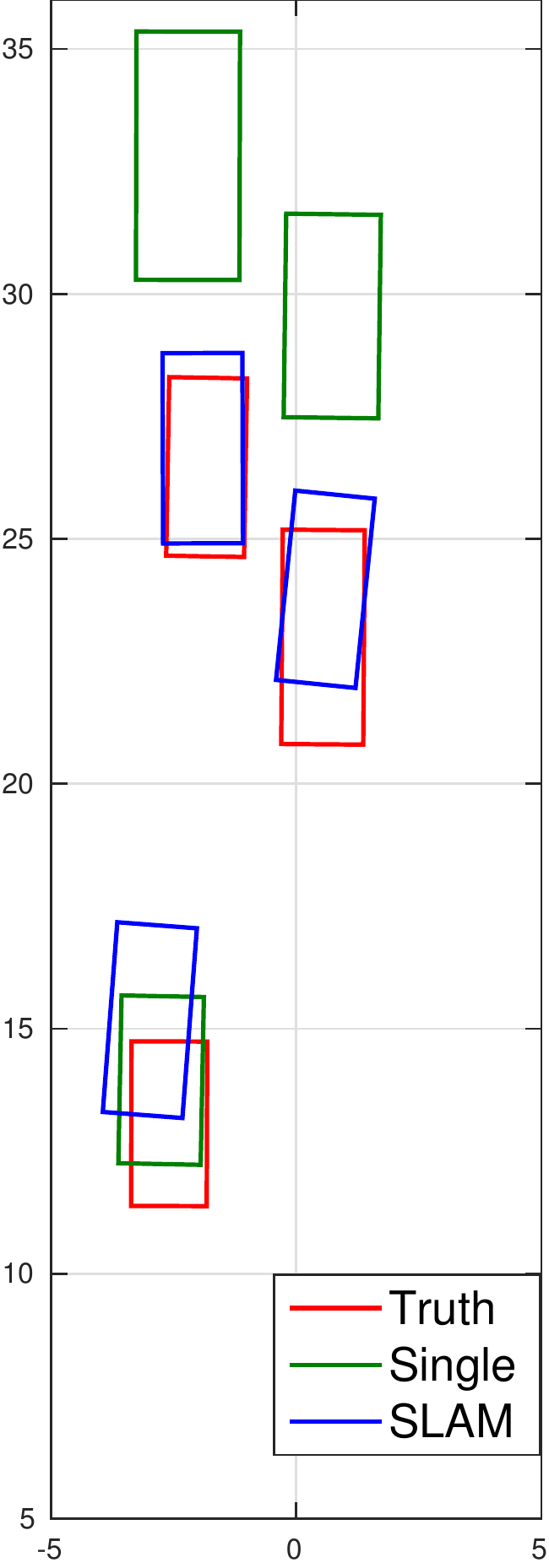}     \label{fig:top view dynamic}}
   \subfigure[]{ \includegraphics[scale=0.15]{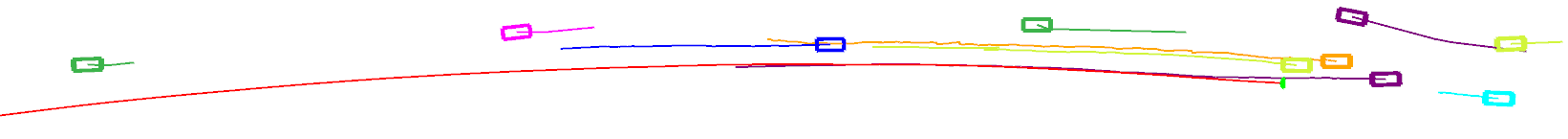}    \label{fig:dynamic history pose}}
   \caption{ Dynamic object SLAM result on KITTI. (a) Sample frames of single image cuboid detection. (b) Top view comparison of 3D detections in single image and multi-view SLAM. (c) Camera and object pose estimation. The red curve starting from left is the camera's trajectory. Other curves attached with rectangle markers represent the dynamic object's trajectory.}
\end{figure}

\begin{figure}[t]
  \centering
   \includegraphics[scale=0.46]{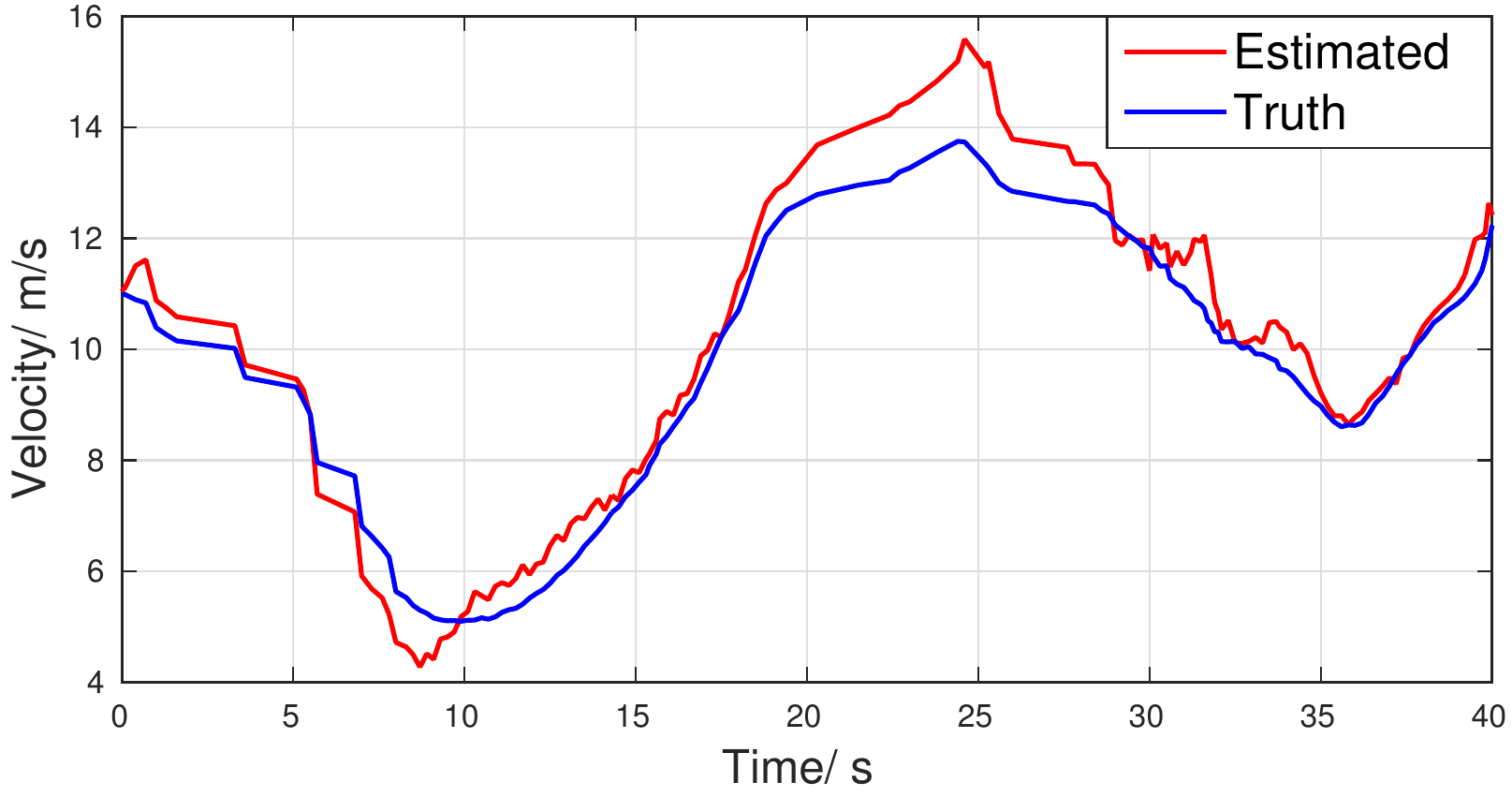}
   \caption{ Dynamic car velocity estimation on KITTI Seq 0047. Our SLAM algorithm based on the piecewise constant motion assumption can correctly estimate the moving object's velocity using just a monocular camera.
   }
   \label{fig:dynamic velocity esti}
\end{figure}

\begin{table*}[t]
\caption{Dynamic Object Detection and Camera Pose Estimation on KITTI Raw Sequence}
\begin{center}
\begin{tabular}{c | c | c c c c c c | c}
\hline
\multicolumn{2}{c|}{Seq}   &13  &14  &15  &04  &56  &32  &Mean \\ \hline
Object  &Single view  &0.41  &0.11   &0.42  &0.11   &\textbf{0.54} &\textbf{0.34}  &0.32\\
3D IoU   &Ours 	  &\textbf{0.51}   &\textbf{0.28}   &\textbf{0.44}  & \textbf{0.42}   &0.42  &0.26   &\textbf{0.39} \\ \hline
Trans  &ORB-No LC   &2.34   &11.5   &2.4  &2.31   &8.45  &\textbf{2.76}  &4.96\\
error(m)    &Ours 	 &\textbf{0.99}   &\textbf{7.62}  &\textbf{1.94}   &\textbf{1.50}   &\textbf{5.39}  &3.07 &\textbf{3.42}\\ \hline
\end{tabular}
\end{center}
\label{table:dynamic kitti error}
\end{table*}

\begin{table*}[t]
\caption{Dynamic Object Localization Comparison on KITTI Raw Sequence}
\begin{center}
\begin{tabular}{c | c | c c c c | c c c c | c | c}
\hline
\multicolumn{2}{c|}{Seq}   	&\multicolumn{4}{c|}{04}  &\multicolumn{4}{c|}{47}  &56  &\multirow{3}{*}{Mean} \\ \cline{1-11}
\multicolumn{2}{c|}{Obj. ID} &1    &2    &3    &6  	  &0    &4    &9    &12 	   &0    \\ \cline{1-11}
\multicolumn{2}{c|}{No. Frames}  &91   &251  &284  &169  	  &170  &96   &94   &637	   &293  \\ \hline
Depth &\cite{song2014robust} &\textbf{4.1}  &6.8	  &5.3  &7.3   &9.6	&11.4   &7.1  & 10.5  &5.5  &7.9 \\
error &\cite{song2015joint}  &6.0  &\textbf{5.6}  &4.9  &5.9  &\textbf{5.9}  &12.5  &7.0  &8.2  &6.0  &6.8 \\
(\%)  &Ours  			&5.1  &11.1  &\textbf{1.8}  &\textbf{2.5}  &9.3  &\textbf{3.8}  &\textbf{6.6}  &\textbf{2.8}  &\textbf{0.8}  &\textbf{4.9} \\ \hline
\end{tabular}
\end{center}
\label{table:dynamic obj detect comp}
\end{table*}

\subsection{Dynamic object}
We also test the algorithm on the dynamic car sequences on the KITTI datasets as shown in Fig. \ref{fig:dynamic sample image}. We select some raw sequences with more dynamic objects observed over a long time shown in Table \ref{table:dynamic kitti error}. The full name of the sequences is ``2011\_0926\_00xx''. The first four sequences also correspond to Seq $3,4,5,18$ in KITTI tracking dataset. Most cars in these sequences are moving. Ground truth object annotations are available for all or some frames and ground truth camera poses are also provided by GPS/INS. 


\subsubsection{Qualitative results}

Some single image detection examples are shown in Fig. \ref{fig:dynamic sample image}. Fig. \ref{fig:top view dynamic} shows the top view of the first image of Fig. \ref{fig:dynamic sample image}. For the two distant front cars, even though the 2D image cuboid detection looks good, it actually has a large 3D distance error. This is because only the car's back faces are observable causing ill-constraint single image detection. After multi-view dynamic object BA, the blue optimized object matches better with the ground truth mostly due to the motion model constraints. However, it can sometimes decrease the accuracy for example the bottom object in the figure. Some possible reasons are due to the noisy 2D and 3D object detection, especially for the close objects. The constant motion assumption may also cause errors when the vehicle accelerates or decelerates. Fig. \ref{fig:dynamic history pose} shows all the dynamic objects' history poses as well as camera poses. The objects' trajectories are smooth due to the motion model constraints.

Fig. \ref{fig:dynamic velocity esti} shows the velocity estimation of one of objects in on Seq 0047 data. We can see that the computed ground truth object velocity also changes with time, therefore, the piecewise constant velocity motion explained in Sec. \ref{sec: dynamic implementation} is reasonable. With a monocular camera, the proposed algorithm can roughly estimate the object's absolute velocity.

\subsubsection{Quantitative results}

Since there is no current monocular SLAM utilizing dynamic objects to change camera pose estimation, we thus directly compare with the state-of-the-art featured based ORB SLAM. Though it already has some modules to detect dynamic points as outliers based on reprojection error, to compare with ORB SLAM fairly, we directly remove features lying in the dynamic object areas and report its pose estimation result. From Table \ref{table:dynamic kitti error}, we can see our method can improve the camera pose estimation on most sequences especially when objects can be observed and tracked over many consecutive frames for example in the first four datasets. This is because with more observations, the objects' velocity and dynamic points' position can be estimated more accurately and thus have more effect on the camera pose estimation, while in the last two sequences, objects are usually observed by only a few frames.

We also compare the 3D object localization with other monocular methods shown in Table \ref{table:dynamic obj detect comp}. The most similar one to us is \cite{song2015joint} which utilized semantic and geometric costs to optimize object locations, but their approach assumed the camera poses are already solved and fixed. We utilize the same metric in \cite{song2015joint} to measure the relative object depth error in each frame and the number is taken directly from the paper. From the table, our method outperforms others on most sequences. In the sequence 56, the relative depth error is only $0.8\%$.

\begin{table}[t]
\caption[Average Runtime of Different System Components]{Runtime of Different System Components}
\begin{center}
\begin{tabular}{c |  c | c c}
\hline
\multirow{2}{*}{Dataset}  & \multirow{2}{*}{Tasks} 	&Runtime \\ 
						  &							& (mSec) \\ \hline
				&Tracking thread (per frame)  		&33.0 \\
Outdoor  		&No object BA   			&182.7 \\ 
KITTI      	&Static object BA    		&194.5 \\ 
		  		&Dynamic object BA   		&365.2 \\  \hline
\multirow{3}{*}{\shortstack{Indoor\\ ICL room}}   		&Tracking thread  (per frame) 		&15.0   \\ 
 	  	&No object BA   			&49.5   \\  
  				&Static object BA   	    &55.3   \\ \hline

\end{tabular}
\end{center}
\label{table:time analysis 2}
\end{table}

\subsection{Time Analysis}
\label{sec: time analysis}
Finally, we provide the computational analysis of our system. The experiments are carried out on Intel i7-4790 CPU at 4.0 GHz. The 2D object detection time depends on the GPU power and CNN model complexity. Many algorithms such as Yolo can run in real time. After getting the 2D boxes, our 3D cuboid detection takes about 20ms per image and the main computation of it is edge detection. For the SLAM part implemented in C++ on CPU, we show the time usage in Table \ref{table:time analysis 2}. The computation strongly depends on the datasets with different image size and textures, therefore we choose two representative sequences: outdoor KITTI 07 at 10Hz shown in Fig. \ref{fig:kitti mapping} and indoor ICL-NUIM livingroom dataset at 30Hz shown in Fig. \ref{fig:icl mapping}. On average, there are 5 object landmarks in each local BA optimization in the two datasets. The tracking thread includes the ORB point feature detection and camera pose tracking for each frame which can run in real time from the table. The bundle adjustment (BA) map optimization occurs when a new keyframe is created, therefore it does not need to run in real-time. We show the time usage of BA when different types of landmarks exist. In the static environment, adding objects into the system only increases the optimization by $7\%$. This is reasonable as there are only a few objects in the local map optimization. For the dynamic cases, due to many new variables and measurements of dynamic points, the computation increases by a factor of two.

\section{Conclusion}
\label{sec:conclusion}
In this work, we convey a general approach for monocular 3D object detection and SLAM mapping without prior object models, in both static and dynamic environments. More importantly, we demonstrate for the first time, that semantic object detection and geometric SLAM can benefit each other in one unified framework.

For the single image 3D object detection, we propose a new method to efficiently generate high quality cuboid proposals from the 2D bounding box based on vanishing points. Proposals are then scored efficiently by image cues. In the SLAM part, we propose an object level SLAM with novel measurement functions between cameras, objects and points, and new object association methods to robustly handle occlusion and dynamic movement. Objects can provide long range geometric and scale constraints for camera pose estimation. In turn, SLAM also provides camera pose initialization for detecting and refining 3D object. For the dynamic scenarios, we also show that with the new measurement constraints, the moving object and point can also improve the camera pose estimation through the tightly coupled optimization.

We evaluate the two parts on different indoor and outdoor datasets and achieve the best accuracy of 3D object detection on SUN RGBD subset data and camera pose estimation on KITTI odometry datasets. In the future, we are also interested in the dense mapping using objects. More complete scene understanding can also be integrated with SLAM optimization.





\bibliographystyle{unsrt}    
\bibliography{ref}

\begin{thebibliography}{10}

\bibitem{he2017mask}
Kaiming He, Georgia Gkioxari, Piotr Doll{\'a}r, and Ross Girshick.
\newblock Mask {R-CNN}.
\newblock {\em IEEE International Conference on Computer Vision (ICCV)}, 2017.

\bibitem{xiang2017subcategory}
Yu~Xiang, Wongun Choi, Yuanqing Lin, and Silvio Savarese.
\newblock Subcategory-aware convolutional neural networks for object proposals
  and detection.
\newblock In {\em Applications of Computer Vision (WACV), 2017 IEEE Winter
  Conference on}, pages 924--933. IEEE, 2017.

\bibitem{chabot2017deep}
Florian Chabot, Mohamed Chaouch, Jaonary Rabarisoa, C{\'e}line Teuli{\`e}re,
  and Thierry Chateau.
\newblock Deep {MANTA}: A coarse-to-fine many-task network for joint 2d and
  {3D} vehicle analysis from monocular image.
\newblock In {\em IEEE Conference on Computer Vision and Pattern Recognition
  (CVPR)}, pages 2040--2049, 2017.

\bibitem{mur2015orb}
Raul Mur-Artal, JMM Montiel, and Juan~D Tardos.
\newblock {ORB-SLAM}: a versatile and accurate monocular {SLAM} system.
\newblock {\em IEEE Transactions on Robotics}, 31(5):1147--1163, 2015.

\bibitem{chao2017edge}
Shichao Yang and Sebastian Scherer.
\newblock Direct monocular odometry using points and lines.
\newblock In {\em IEEE international conference on Robotics and automation
  (ICRA)}. IEEE, 2017.

\bibitem{kaess2015simultaneous}
Michael Kaess.
\newblock Simultaneous localization and mapping with infinite planes.
\newblock In {\em International Conference on Robotics and Automation (ICRA)},
  pages 4605--4611. IEEE, 2015.

\bibitem{murthy2017reconstructing}
J~Krishna Murthy, GV~Sai Krishna, Falak Chhaya, and K~Madhava Krishna.
\newblock Reconstructing vehicles from a single image: Shape priors for road
  scene understanding.
\newblock In {\em Robotics and Automation (ICRA), 2017 IEEE International
  Conference on}, pages 724--731. IEEE, 2017.

\bibitem{lim2013parsing}
Joseph~J Lim, Hamed Pirsiavash, and Antonio Torralba.
\newblock Parsing {IKEA} objects: Fine pose estimation.
\newblock In {\em IEEE International Conference on Computer Vision}, pages
  2992--2999, 2013.

\bibitem{kehl2017ssd}
Wadim Kehl, Fabian Manhardt, Federico Tombari, Slobodan Ilic, and Nassir Navab.
\newblock {SSD-6D}: Making rgb-based {3D} detection and 6d pose estimation
  great again.
\newblock In {\em IEEE International Conference on Computer Vision (ICCV)},
  2017.

\bibitem{xiao2012localizing}
Jianxiong Xiao, Bryan Russell, and Antonio Torralba.
\newblock Localizing {3D} cuboids in single-view images.
\newblock In {\em Advances in neural information processing systems (NIPS)},
  pages 746--754, 2012.

\bibitem{hedau2010thinking}
Varsha Hedau, Derek Hoiem, and David Forsyth.
\newblock Thinking inside the box: Using appearance models and context based on
  room geometry.
\newblock In {\em European Conference on Computer Vision}, pages 224--237.
  Springer, 2010.

\bibitem{chen2016monocular}
Xiaozhi Chen, Kaustav Kundu, Ziyu Zhang, Huimin Ma, Sanja Fidler, and Raquel
  Urtasun.
\newblock Monocular {3D} object detection for autonomous driving.
\newblock In {\em IEEE Conference on Computer Vision and Pattern Recognition
  (CVPR)}, pages 2147--2156, 2016.

\bibitem{mousavian20163d}
Arsalan Mousavian, Dragomir Anguelov, John Flynn, and Jana Kosecka.
\newblock {3D} bounding box estimation using deep learning and geometry.
\newblock {\em IEEE Conference on Computer Vision and Pattern Recognition
  (CVPR)}, 2017.

\bibitem{li2018stereo}
Peiliang Li, Tong Qin, and andShaojie Shen.
\newblock Stereo vision-based semantic {3D} object and ego-motion tracking for
  autonomous driving.
\newblock In {\em The European Conference on Computer Vision (ECCV)}, September
  2018.

\bibitem{engel2017direct}
Jakob Engel, Vladlen Koltun, and Daniel Cremers.
\newblock Direct sparse odometry.
\newblock {\em IEEE Transactions on Pattern Analysis and Machine Intelligence},
  2017.

\bibitem{dong2017visual}
Jingming Dong, Xiaohan Fei, and Stefano Soatto.
\newblock Visual-inertial-semantic scene representation for {3D} object
  detection.
\newblock In {\em IEEE Conference on Computer Vision and Pattern Recognition
  (CVPR)}, 2017.

\bibitem{pillai2015monocular}
Sudeep Pillai and John Leonard.
\newblock Monocular {SLAM} supported object recognition.
\newblock {\em Robotics: Science and systems}, 2015.

\bibitem{dame2013dense}
Amaury Dame, Victor~A Prisacariu, Carl~Y Ren, and Ian Reid.
\newblock Dense reconstruction using 3d object shape priors.
\newblock In {\em Proceedings of the IEEE Conference on Computer Vision and
  Pattern Recognition}, pages 1288--1295, 2013.

\bibitem{bao2012semantic}
Sid~Yingze Bao, Mohit Bagra, Yu-Wei Chao, and Silvio Savarese.
\newblock Semantic structure from motion with points, regions, and objects.
\newblock In {\em Conference on Computer Vision and Pattern Recognition
  (CVPR)}, pages 2703--2710. IEEE, 2012.

\bibitem{salas2013slam++}
Renato~F Salas-Moreno, Richard~A Newcombe, Hauke Strasdat, Paul~HJ Kelly, and
  Andrew~J Davison.
\newblock {SLAM+}: Simultaneous localisation and mapping at the level of
  objects.
\newblock In {\em IEEE Conference on Computer Vision and Pattern Recognition},
  pages 1352--1359, 2013.

\bibitem{frost2018recovering}
Duncan Frost, Victor Prisacariu, and David Murray.
\newblock Recovering stable scale in monocular slam using object-supplemented
  bundle adjustment.
\newblock {\em IEEE Transactions on Robotics}, 34(3):736--747, 2018.

\bibitem{sucar2017bayesian}
Edgar Sucar and Jean-Bernard Hayet.
\newblock Bayesian scale estimation for monocular {SLAM} based on generic
  object detection for correcting scale drift.
\newblock {\em IEEE International Conference on Robotics and Automation
  (ICRA)}, 2018.

\bibitem{galvez2016real}
Dorian G{\'a}lvez-L{\'o}pez, Marta Salas, Juan~D Tard{\'o}s, and JMM Montiel.
\newblock Real-time monocular object {SLAM}.
\newblock {\em Robotics and Autonomous Systems}, 75:435--449, 2016.

\bibitem{rubino20183d}
Cosimo Rubino, Marco Crocco, and Alessio Del~Bue.
\newblock {3D} object localisation from multi-view image detections.
\newblock {\em IEEE transactions on pattern analysis and machine intelligence},
  40(6):1281--1294, 2018.

\bibitem{nicholson2018quadricslam}
Lachlan~James Nicholson, Michael~J Milford, and Niko Sunderhauf.
\newblock {QuadricSLAM}: Dual quadrics from object detections as landmarks in
  object-oriented {SLAM}.
\newblock {\em IEEE Robotics and Automation Letters}, 4(1):1--8, 2019.

\bibitem{bowman2017probabilistic}
Sean~L Bowman, Nikolay Atanasov, Kostas Daniilidis, and George~J Pappas.
\newblock Probabilistic data association for semantic {SLAM}.
\newblock In {\em Robotics and Automation (ICRA), 2017 IEEE International
  Conference on}, pages 1722--1729. IEEE, 2017.

\bibitem{syang2016popslam}
Shichao Yang, Yu~Song, Michael Kaess, and Sebastian Scherer.
\newblock {Pop-up SLAM}: a semantic monocular plane {SLAM} for low-texture
  environments.
\newblock In {\em International conference on Intelligent Robots and Systems
  (IROS)}. IEEE, 2016.

\bibitem{yang2018deep}
Nan Yang, Rui Wang, J{\"o}rg St{\"u}ckler, and Daniel Cremers.
\newblock Deep virtual stereo odometry: Leveraging deep depth prediction for
  monocular direct sparse odometry.
\newblock In {\em European Conference on Computer Vision}, pages 835--852.
  Springer, 2018.

\bibitem{wang2017deepvo}
Sen Wang, Ronald Clark, Hongkai Wen, and Niki Trigoni.
\newblock Deepvo: Towards end-to-end visual odometry with deep recurrent
  convolutional neural networks.
\newblock In {\em Robotics and Automation (ICRA), 2017 IEEE International
  Conference on}, pages 2043--2050. IEEE, 2017.

\bibitem{barsan2018robust}
Ioan~Andrei B{\^a}rsan, Peidong Liu, Marc Pollefeys, and Andreas Geiger.
\newblock Robust dense mapping for large-scale dynamic environments.
\newblock In {\em IEEE International Conference on Robotics and Automation
  (ICRA)}, 2018.

\bibitem{bescos2018dynslam}
Berta Besc{\'o}s, Jos{\'e}~M F{\'a}cil, Javier Civera, and Jos{\'e} Neira.
\newblock {DynSLAM}: Tracking, mapping and inpainting in dynamic scenes.
\newblock {\em IEEE Robotics and Automation Letters}, 2018.

\bibitem{song2015joint}
Shiyu Song and Manmohan Chandraker.
\newblock Joint {SFM} and detection cues for monocular {3D} localization in
  road scenes.
\newblock In {\em IEEE Conference on Computer Vision and Pattern Recognition
  (CVPR)}, pages 3734--3742, 2015.

\bibitem{reddy2015dynamic}
N~Dinesh Reddy, Prateek Singhal, Visesh Chari, and K~Madhava Krishna.
\newblock Dynamic body {VSLAM} with semantic constraints.
\newblock In {\em Intelligent Robots and Systems (IROS), 2015 IEEE/RSJ
  International Conference on}, pages 1897--1904. IEEE, 2015.

\bibitem{henein18exploiting}
Mina Henein, Gerard Kennedy, Robert Mahony, and Viorela Ila.
\newblock Exploiting rigid body motion for {SLAM} in dynamic environments.
\newblock In {\em IEEE International Conference on Robotics and Automation
  (ICRA)}, 2018.

\bibitem{hartley2003multiple}
Richard Hartley and Andrew Zisserman.
\newblock {\em Multiple view geometry in computer vision}.
\newblock Cambridge university press, 2003.

\bibitem{von2008lsd}
Rafael~Grompone von Gioi, Jeremie Jakubowicz, Jean-Michel Morel, and Gregory
  Randall.
\newblock {LSD}: A fast line segment detector with a false detection control.
\newblock {\em IEEE Transactions on Pattern Analysis \& Machine Intelligence},
  (4):722--732, 2008.

\bibitem{kummerle2011g}
Rainer K{\"u}mmerle, Giorgio Grisetti, Hauke Strasdat, Kurt Konolige, and
  Wolfram Burgard.
\newblock g2o: A general framework for graph optimization.
\newblock In {\em Robotics and Automation (ICRA), IEEE International Conference
  on}, pages 3607--3613. IEEE, 2011.

\bibitem{kaess2008isam}
Michael Kaess, Ananth Ranganathan, and Frank Dellaert.
\newblock {iSAM}: Incremental smoothing and mapping.
\newblock {\em Robotics, IEEE Transactions on}, 24(6):1365--1378, 2008.

\bibitem{lavalle2006planning}
Steven~M LaValle.
\newblock {\em Planning algorithms}.
\newblock Cambridge university press, 2006.

\bibitem{henriques2015high}
Jo{\~a}o~F Henriques, Rui Caseiro, Pedro Martins, and Jorge Batista.
\newblock High-speed tracking with kernelized correlation filters.
\newblock {\em IEEE Transactions on Pattern Analysis and Machine Intelligence},
  37(3):583--596, 2015.

\bibitem{redmon2016yolo9000}
Joseph Redmon and Ali Farhadi.
\newblock {YOLO9000}: better, faster, stronger.
\newblock {\em Computer Vision and Pattern Recognition (CVPR)}, 2017.

\bibitem{cai16mscnn}
Zhaowei Cai, Quanfu Fan, Rogerio Feris, and Nuno Vasconcelos.
\newblock A unified multi-scale deep convolutional neural network for fast
  object detection.
\newblock In {\em ECCV}, 2016.

\bibitem{song2015sun}
Shuran Song, Samuel~P Lichtenberg, and Jianxiong Xiao.
\newblock {SUN RGB-D}: A {RGB-D} scene understanding benchmark suite.
\newblock In {\em IEEE Conference on Computer Vision and Pattern Recognition
  (CVPR)}, pages 567--576, 2015.

\bibitem{geiger2012we}
Andreas Geiger, Philip Lenz, and Raquel Urtasun.
\newblock Are we ready for autonomous driving? the kitti vision benchmark
  suite.
\newblock In {\em Computer Vision and Pattern Recognition (CVPR), 2012 IEEE
  Conference on}, pages 3354--3361. IEEE, 2012.

\bibitem{chen20153d}
Xiaozhi Chen, Kaustav Kundu, Yukun Zhu, Andrew~G Berneshawi, Huimin Ma, Sanja
  Fidler, and Raquel Urtasun.
\newblock {3D} object proposals for accurate object class detection.
\newblock In {\em Advances in Neural Information Processing Systems}, pages
  424--432, 2015.

\bibitem{choi2013understanding}
Wongun Choi, Yu-Wei Chao, Caroline Pantofaru, and Silvio Savarese.
\newblock Understanding indoor scenes using {3D} geometric phrases.
\newblock In {\em IEEE Conference on Computer Vision and Pattern Recognition},
  pages 33--40, 2013.

\bibitem{sturm2012benchmark}
J{\"u}rgen Sturm, Nikolas Engelhard, Felix Endres, Wolfram Burgard, and Daniel
  Cremers.
\newblock A benchmark for the evaluation of {RGB-D} {SLAM} systems.
\newblock In {\em Intelligent Robots and Systems (IROS), IEEE/RSJ International
  Conference on}, pages 573--580. IEEE, 2012.

\bibitem{handa:etal:ICRA2014}
A.~Handa, T.~Whelan, J.B. McDonald, and A.J. Davison.
\newblock A benchmark for {RGB-D} visual odometry, {3D} reconstruction and
  {SLAM}.
\newblock In {\em IEEE Intl. Conf. on Robotics and Automation, ICRA}, Hong
  Kong, China, May 2014.

\bibitem{lee2015online}
Bhoram Lee, Kostas Daniilidis, and Daniel~D Lee.
\newblock Online self-supervised monocular visual odometry for ground vehicles.
\newblock In {\em Robotics and Automation (ICRA), 2015 IEEE International
  Conference on}, pages 5232--5238. IEEE, 2015.

\bibitem{song2016high}
Shiyu Song, Manmohan Chandraker, and Clark~C Guest.
\newblock High accuracy monocular {SFM} and scale correction for autonomous
  driving.
\newblock {\em IEEE transactions on pattern analysis and machine intelligence},
  38(4):730--743, 2016.

\bibitem{song2014robust}
Shiyu Song and Manmohan Chandraker.
\newblock Robust scale estimation in real-time monocular {SFM} for autonomous
  driving.
\newblock In {\em Proceedings of the IEEE Conference on Computer Vision and
  Pattern Recognition}, pages 1566--1573, 2014.

\end{thebibliography}

\addtolength{\textheight}{-12cm}   

\end{document}